\documentclass[journal]{IEEEtran}
\ifCLASSINFOpdf
\else
\fi

\usepackage[square,comma,numbers,sort&compress]{natbib}
\usepackage{color}
\usepackage{graphicx}
\usepackage{amsmath}
\usepackage{algorithm}
\usepackage{algorithmic}
\usepackage{bm}
\usepackage{amssymb}
\usepackage{epstopdf}
\usepackage{booktabs}
\usepackage{subfigure}
\usepackage{stfloats}
\usepackage{multirow}
\usepackage{amsthm}

\begin{document}
\title{Depthwise Convolution for Multi-Agent Communication with  Enhanced Mean-Field Approximation}

\author{Donghan~Xie, Zhi~Wang,~\IEEEmembership{Member,~IEEE,}
	        Chunlin~Chen,~\IEEEmembership{Senior Member,~IEEE},   and~Daoyi~Dong,~\IEEEmembership{Fellow,~IEEE}%

\thanks{This work is published on \textit{IEEE Transactions on Neural Networks and Learning Systems}, 2022, DOI: 10.1109/TNNLS.2022.3230701.}

\thanks{This work was supported in part by the National Natural Science Foundation of China under Grant 62006111 and Grant 62073160, in part by the Australian Research Council's Future Fellowship funding scheme under Project FT220100656, in part by the Natural Science Foundation of Jiangsu Province of China under Grant BK20200330, and in part by the Alexander von Humboldt Foundation, Germany. \textit{(Corresponding author: Zhi Wang.)}}

\thanks{Donghan Xie is with the Department of Control Science and Intelligence Engineering, School of Management and Engineering, Nanjing University, Nanjing 210093, China, and is also with Tencent, Shenzhen 518000, China (e-mail: donghanxie@smail.nju.edu.cn).} 

\thanks{Zhi Wang and Chunlin Chen are with the Department of Control Science and Intelligence Engineering, School of Management and Engineering, Nanjing University, Nanjing 210093, China (e-mail: zhiwang@nju.edu.cn; clchen@nju.edu.cn).}% <-this % stops a space
\thanks{Daoyi Dong is with the School of Engineering and Information Technology, University of New South Wales, Canberra, ACT 2600, Australia (e-mail: daoyidong@gmail.com).}% <-this % stops a space
}

\maketitle

\begin{abstract}
Multi-agent settings remain a fundamental challenge in the reinforcement learning (RL) domain due to the partial observability and the lack of accurate real-time interactions across agents.
In this paper, we propose a new method based on local communication learning to tackle the multi-agent RL (MARL) challenge within a large number of agents coexisting.
First, we design a new communication protocol that exploits the ability of depthwise convolution to efficiently extract local relations and learn local communication between neighboring agents.
To facilitate multi-agent coordination, we explicitly learn the effect of joint actions by taking the policies of neighboring agents as inputs.
Second, we introduce the mean-field approximation into our method to reduce the scale of agent interactions.
To more effectively coordinate behaviors of neighboring agents, we enhance the mean-field approximation by a supervised policy rectification network (PRN) for rectifying real-time agent interactions and by a learnable compensation term for correcting the approximation bias.
The proposed method enables efficient coordination as well as outperforms several baseline approaches on the adaptive traffic signal control (ATSC) task and the StarCraft II multi-agent challenge (SMAC).
\end{abstract}

\begin{IEEEkeywords}
Agent communication, depthwise convolution, mean-field approximation, multi-agent reinforcement learning, StarCraft II multi-agent challenge.
\end{IEEEkeywords}

\section{Introduction}
% ======= Introduce MARL ==================================
Based on the Markov decision process (MDP) formulation, reinforcement learning (RL)~\cite{rlai,qlearning} allows agents to solve tasks from direct interactions with the environment, forming an optimal policy to make sequential decisions in a trial-and-error manner~\cite{li2020quantum,zheng2021efficient,li2020deep,wang2019tmechl}.
The recent combination of RL with deep learning, referred to as deep reinforcement learning (DRL)~\cite{dqn}, has emerged as a promising direction for the autonomous acquisition of complex behaviors~\cite{DDPG,PPO}, since it can acquire elaborate skills using general-purpose neural network representations from high-dimensional sensory inputs~\cite{wang2022lifelong,multisource,luo2019balancing,wang2022dirichlet}.
Many artificial intelligence (AI) applications require the collaboration of multiple agents~\cite{MAAC,MAGA,DGN}, and successfully scaling RL to multi-agent settings is crucial to building intelligent systems that can productively interact with each other and humans.

Multi-agent reinforcement learning (MARL) is concerned with coordinating a set of agents toward maximizing each agent's or the group's objective, where each individual can only observe a local part of the shared environment~\cite{vinyals2019grandmaster}.
A fundamental challenge in MARL is to tackle the non-stationarity due to the partial observability and the lack of accurate real-time interactions across agents~\cite{i2c,yumacar}.
Fully centralized control that unifies all agents into a single one is usually infeasible due to the exponential growth of the size of joint action spaces.
% The simplest option to overcome the curse of dimensionality is to learn an individual action-value function independently for each agent, as in independent Q-learning (IQL)~\cite{iql,tampuu2017multiagent}, while the learning is often unstable as changes in one agent's policy will affect those of the others.
The simplest option to overcome the curse of dimensionality is to learn an individual action-value function independently for each agent, as in independent Q-learning (IQL)~\cite{iql}, while the learning is often unstable as changes in one agent's policy will affect those of the others.
This issue can be mitigated by the centralized training and decentralized execution (CTDE) paradigm~\cite{maddpg,foerster2018counterfactual} that typically leverages centralized critics to approximate the global value function of the joint policy and trains actors restricted to the local observation of a single agent~\cite{vdn,ebgcn,rashid2018qmix,rashid2020weighted,gupta21uneven,zhang21fop}.
However, the execution phase can still suffer from non-stationarity due to not accounting for extra information from other agents.

% ======= Introduce Communication ==========================
Due to the partial observability and limited channel capacity, a communication protocol is vital to coordinate the behavior of agents and solve the task via information sharing~\cite{commnet,tarmac,lv2022logic}.
A straightforward approach is to learn global communication that shares information across all agents, such as DIAL~\cite{dial}, BiCNet~\cite{bicnet}, and IMAC~\cite{imac}.
The computational cost can be high for all agents communicating with each other, especially with a large number of agents coexisting.
Instead, the other kind of approaches attempts to learn informative local communication that needs to efficiently exploit the available communication resources, such as ATOC~\cite{atoc}, IC3Net~\cite{ic3net}, and I2C~\cite{i2c}. 
While these methods exploit extra information from neighboring agents, they do not explicitly learn the effect of joint actions.
It could potentially reduce the coordination efficiency since the environment dynamics of an agent depends on actions of the others.

In this paper, we propose a new MARL method within a large number of agents coexisting based on local communication learning.
First, inspired by the fact that convolution is widely used in extracting local relations~\cite{lecun2015deep}, we design a new communication protocol that exploits the ability of depthwise convolution~\cite{howard2017mobilenets} to efficiently learn local communication between neighboring agents.
Since the environment dynamics of an agent depends on actions of the others, we take the policies of neighboring agents as inputs and explicitly learn the effect of joint actions to facilitate multi-agent coordination.
Second, in our method, the mean-field approximation~\cite{mean} is introduced to approximate the interactions within the population of agents considering the average effect from the neighboring agents of an individual, thus considerably reducing the scale of agent interactions.
To more accurately coordinate behaviors of neighboring agents, we enhance the mean-field approximation by a supervised policy rectification network (PRN) that predicts real-time policies from previous information to rectify real-time agent interactions.
\footnote{The communication of real-time information during execution may lead to a certain time lag due to bandwidth limitations in real-world applications~\citep{tarmac}. 
Hence, throughout the paper, we assume that an individual cannot know the real-time information of its neighbors during execution.}
Besides, since mean-field approximation drops out the second-order remainders, we attempt through learning to compensate for the approximation bias via our communication protocol and obtain a more accurate mean-field estimate of agent interactions.

We extensively evaluate our method on two multi-agent tasks: the adaptive traffic signal control (ATSC) on the SUMO platform~\cite{SUMO,yumacar} and the StarCraft II multi-agent challenge (SMAC)~\cite{smac}.
Experimental results show that our method can achieve more efficient multi-agent coordination and outperform several baseline approaches.

In summary, our contributions are threefold:
\begin{enumerate}
	\item We propose a new protocol that exploits the ability of depthwise convolution to efficiently learn local communication, and we explicitly learn the effect of joint actions to facilitate multi-agent coordination. 
	\item We exploit the mean-field approximation to reduce the scale of agent interactions, and enhance the mean-field estimate by a supervised policy rectification network (PRN) and a learnable compensation term.
	\item We perform extensive experiments to verify that our method can consistently improve the multi-agent learning performance over several baselines.
\end{enumerate}

The remainder of this paper is organized as follows.
Section~\ref{related} gives the related work on MARL.
Section~\ref{prelimilaries} introduces preliminaries of MARL and mean-field approximation.
Section~\ref{method} first presents the proposed DCCP and enhanced mean-field approximation, followed by the final integrated algorithm.
Experiments on the ATSC and SMAC tasks are conducted in Section~\ref{experiments}.
Section~\ref{conclusions} presents concluding remarks.

\section{Related Work}\label{related}
Communication learning is recognized as a promising way to handle multi-agent systems by sharing extra information (e.g., observations and policies) with each other for coordination~\citep{jin2022event2,lv2022separation,jin2020event,lv2022consensus}.
A straightforward approach is global information sharing among all agents.
\citet{dial} proposed to learn one-round point-to-point communication, which uses the broadcast messages from the previous time step instead of real-time ones due to communication constraints in the real world.
\citet{bicnet} used the bi-directional RNN to maintain the communication protocol that can learn various types of coordination strategies.
IMAC~\citep{imac} learned an efficient protocol that compresses communication messages and schedules more accurate information delivery to overcome the bandwidth limitations.
Some recent works, such as TarMAC \cite{tarmac}, SARNet \cite{sar}, and DICG~\cite{dicg}, employed an attention mechanism with global communication to learn what messages to send and whom to address these messages to.
\citet{ndq} designed an expressive and succinct communication protocol by introducing information-theoretic regularizers for maximizing mutual information between agents' action selection and communication messages.
\citet{jin2022event} utilized the event-triggered mechanism to reduce a large amount of consumption of continuous communication at the cost of a small amount of computing resources.
The computational cost can be high for all agents transmitting a large amount of information to each other, especially with many agents coexisting.

Instead, the other kind of approaches attempts to learn local communication that needs efficiently exploiting the available communication resources.
\citet{commnet} extended CommNet to a local variant that allows agents to communicate to others within a certain range only.
\citet{atoc} proposed an attentional model that dynamically determines whether the agent should communicate with other agents to cooperate in its observable field.
\citet{ic3net} learned when to communicate by a gating mechanism with individualized rewards to gain better performance and scalability.
\citet{i2c} turned to realize peer-to-peer communication by using causal inference to learn a prior network that maps the agent's local observation to a belief about whom to communicate with. 
While these local communication learning methods account for the extra information from other agents, they do not explicitly learn the effect of joint actions, which could potentially reduce coordination efficiency since the environment dynamics of an agent depends on the others.

Another thread of work is to consider the extra information by explicitly learning policies of other agents.
\citet{hyperq} proposed hyper Q-learning that estimates policies of other agents using Bayesian inference, which is only feasible for discrete or low-dimensional tasks due to the high computational burden of computing exact Bayesian posteriors.
\citet{finger} extended hyper Q-learning to high-dimensional and continuous state spaces by communicating only two scalars (i.e., the current learning step and the learning rate) among agents, while these two scalars have limited representation power for conjecturing the underlying policies.
\citet{mean} introduced the mean-field theory~\cite{meantheory} into MARL to reduce the scale of agent interactions, by using interactions only between an individual and the average effect from its neighboring agents to approximate those interactions within a population of agents.

Here, we exploit the ability of depthwise convolution to efficiently extract local relations and learn local communication between neighboring agents.
In contrast to existing local communication learning approaches, we take the policies of neighboring agents as inputs and explicitly learn the effect of joint actions to facilitate coordination efficiency.
Further, we enhance the mean-field approximation with a supervised PRN and a learnable compensation term to obtain a more accurate mean-field estimate of agent interactions.

\section{Preliminaries}\label{prelimilaries}
% \subsection{Multi-agent Reinforcement Learning}
\subsection{Reinforcement Learning (RL)}
% Reinforcement learning is studied to deal with the sequencial decision-making problems through the Markov decision process (MDP) framework.
RL is studied to deal with the sequential decision-making problems through the Markov decision process (MDP) framework. 
% An MDP is a $5$-tuple $\left(\mathcal{S},\mathcal{A},{T},{R},\gamma\right)$, where $\mathcal{S}$ is the set of states, $\mathcal{A}$ is the set of actions, $T:\mathcal{S}\times\mathcal{A}\times\mathcal{S}\rightarrow[0,1]$ is the conditional transition probabilities between states under the specific action of the environment, $R:\mathcal{S}\times\mathcal{A}\times\mathcal{S}\rightarrow\mathbb{R}$ is the reward function conditioned on current state and action and $\gamma\in[0,1)$ is the discount factor. 
An MDP is concerned with the tuple $\left(\mathcal{S},\mathcal{A},\mathcal{T},\mathcal{R},\gamma\right)$, where $\mathcal{S}$ is the set of states, $\mathcal{A}$ is the set of actions, $\mathcal{T}:\mathcal{S}\times\mathcal{A}\times\mathcal{S}\rightarrow[0,1]$ is the conditional transition probabilities, $\mathcal{R}:\mathcal{S}\times\mathcal{A}\times\mathcal{S}\rightarrow\mathbb{R}$ is the reward function, and $\gamma\in[0,1)$ is the discount factor. 
% We use $\pi(s,a):\mathcal{S}\times\mathcal{A}\rightarrow[0,1]$ to denote a stochastic policy which is the probability of executing action $a$ at state $s$. 
We use $\pi(a|s):\mathcal{S}\times\mathcal{A}\rightarrow[0,1]$ to denote a stochastic policy that is the probability distribution of executing action $a$ at state $s$. 
% The key goal of an agent of reinforcement learning is to find the optimal policy $\pi$ to maximize its expected long-term reward, also called return. We denote the return $J(\pi)$ as:
The goal of RL is to find the optimal policy $\pi^*$ to maximize its expected return $J(\pi)$ as
\begin{equation}
	% J(\pi) = \mathbb{E}_{s_0,a_0,...}\left[\sum_{t=0}^{\infty}\gamma^{t}R(s_t,a_t,s_{t+1})\right],
	J(\pi) = \mathbb{E}_{s_0,a_0,...}\left[\sum\nolimits_{t=0}^{\infty}\gamma^{t}r(s_t,a_t)\right],
\end{equation}
where $a_t\sim \pi(\cdot|s_t)$.

%%%%%%%%%%%%%%%%%%%%%%%%%%%%%%%%%%%%%%%%%%%%%%%%%%%%%%%%%
%%%%%%%%%% 这一段需改成 Deep Q-learning 的介绍 %%%%%%%%%%
%%%%%%%%%%%%%%%%%%%%%%%%%%%%%%%%%%%%%%%%%%%%%%%%%%%%%%%%%
The action value function is defined as the return of policy $\pi$ starting from executing action $a$ in state $s$ as
\begin{equation}
	Q^\pi(s,a) = \mathbb{E}\left[\sum\nolimits_{t=0}^{\infty} \gamma^{t}R(s_t,a_t,s_{t+1})|s_0=s,a_0=a,\pi\right].
\end{equation}
Then, the optimal policy can be directly derived as
\begin{equation}
\pi^*(s) = {\arg\max}_{a\in\mathcal{A}}Q^*(s,a),
\end{equation}
where $Q^*(s,a) = \max_{\pi}Q^\pi(s,a)$.
Deep Q-learning~\citep{dqn} represents the Q-function $Q(s,a;\bm{\theta})$ with a neural network parameterized by $\bm{\theta}$.
During training, the transition tuples $(s,a,r,s^\prime)$ are stored in a \textit{replay buffer}. 
The parameters $\bm{\theta}$ are updated by iteratively sampling a batch of transitions from the buffer and minimizing the squared temporal-difference error as
\begin{equation}
	\mathcal{L}(\bm{\theta}) = \sum\nolimits_{i}\left[\left( y_i^{\text{target}} - Q\left(s,a;\bm{\theta}\right)   \right)^2\right],
\end{equation}
where the target Q-values can be formulated as
\begin{equation}
y_i^{\text{target}} = \left\{
\begin{aligned}
& r_i, && \text{if terminated,}  \\
& r_i+\gamma \max\nolimits_{a'_i}Q(s'_i,a'_i;\bm{\theta}^-), && \text{otherwise,}  
\end{aligned}
\right.
\end{equation}
and $\bm{\theta}^{-}$ are parameters of a target network that is periodically frozen and synchronized from $\bm{\theta}$ for several iterations.

\subsection{Multi-Agent Reinforcement Learning (MARL)}
MARL involves multiple interacting agents coexisting in a sharing environment, where each individual can only observe a local part of the shared environment.
In a multi-agent system, the reward received by an agent depends not only on its own action, but also on the actions taken by the others. 
Through communicating with each other, agents obtain the observations and policies of others to coordinate their behaviors, and learn to maximize an individual or group objective.
Following state-of-the-art works~\citep{rashid2020weighted,MAAC,gupta21uneven}, we formulate the MARL system by a partially observable Markov game~\citep{pomg}, which is a multi-agent extension of Markov decision processes (MDPs).
A partially observable Markov game for $N$ agents is given by a tuple $\left(N,\mathcal{S},\mathcal{A}_1,...,\mathcal{A}_N,T,R_1,...,R_N,O_1,...,O_N,\gamma\right)$, where $\mathcal{S}$ is the state space of the environment, $\mathcal{A}=\mathcal{A}_1\times...\times\mathcal{A}_n$ is the action space, $T$ is the transition function, $R_i:\mathcal{S}\times\mathcal{A}_i\times\mathcal{S}\mapsto \mathbb{R}$ is the reward function of agent $i$, $O_i:\mathcal{S}\times\Omega\mapsto[0,1]$ is a private observation correlated with the state $o_i:\mathcal{S}\mapsto\mathcal{O}_i$, and $\gamma\in[0,1)$ is the discount factor. The objective for each agent $i$ is to maximize its own return $R_i=\sum_{t=0}^{T}\gamma^t r_i^t$.

At each time step, each agent selects its own action $a_i\in\mathcal{A}_i$ conditioned on its own observation $o_i$.
After executing the joint action $\bm{a}=\{a_1,...,a_N\}$, each agent receives its own reward $R^i(\bm{s},\bm{a})$.
Each agent has its own value function $Q^i(\bm{s},\bm{a})$ with respect to the global state $\bm{s}$ and joint action $\bm{a}$ as
\begin{equation}
	Q^i(\bm{s}_{t},\bm{a}_t) = \mathbb{E}_{\bm{s}_{{t+1}:\infty},\bm{a}_{{t+1}:\infty} }\left[\sum\nolimits_{\tau}^{\infty} \gamma^\tau r_{\tau}^i|\bm{s}_t,\bm{a}_t \right].
\end{equation}
This general formalization is a non-cooperative setting, i.e., no explicit coalitions are considered.
When the multi-agent task is fully-cooperative, the reward function is shared among agents as $R^i=R^j (\forall i,j \in \{1,...,N\})$.
Therefore, the value function in cooperative settings can be formalized by a total term $Q^{tot}(\bm{s},\bm{a})$ as in value function factorization approaches~\cite{rashid2018qmix}.

%Since the environment dynamics of a given agent depends on the policies of the others, the learning environment appears non-stationary as agents are independently updating their policies during training. This non-stationarity problem often causes unstable learning processes and brittle convergence properties, which remains a fundamental challenge in MARL. 

\subsection{Mean-Field Approximation}
% Under the hypothesis that global state is availible, mean-field approximation propose to use the interactions between a given agent and a virtual agent whose action is the mean value of the given agent's neighbors, to approximate the value function with respect to joint actions.
Under the hypothesis that the global state is available, mean-field approximation proposes to use the interactions between a given agent and a virtual agent whose action is the mean value of the given agent's neighbors, to approximate the value function with respect to joint actions.
% In MARL, the standard Q-function $Q^j(s,\bm{a})$ is infeasible to learn due to the dimension of joint action $\bm{a}$ exponentially grows with the number of agents.
% In MARL, the standard Q-function $Q^j(s,\bm{a})$ is infeasible to learn since the dimension of joint action $\bm{a}$ exponentially grows with the number of agents.
In MARL, the standard Q-function $Q^i(s,\bm{a})$ is infeasible to learn since the dimension of joint action $\bm{a}$ exponentially grows with the number of agents.
To reduce the complexity of the interactions among agents, mean-field approximation factorizes the Q-function with pairwise local interactions:
\begin{equation}
	Q^i(s,\bm{a}) = \frac{1}{|\mathcal{N}_i|}\sum\nolimits_{j\in \mathcal{N}_i} Q^j(s,a^j,a^j),
	\label{simp}
\end{equation}
where $\mathcal{N}_i$ is the set of the neighbors of agent $i$ with size $|\mathcal{N}_i|$.

The mean action $\bar{a}^i$ is based on the neighborhood $\mathcal{N}_i$, and the action of neighbor $j$ could be expressed by a sum of the mean-action and a small fluctuation $\delta a^{i,j}$ as
\begin{equation}
	a^j = \bar{a}^i+\delta a^{i,j}, \quad \text{where} \quad \bar{a}^i=\frac{1}{|\mathcal{N}_i|}\sum\nolimits_j{a^j}.
	\label{actions}
\end{equation}
Following (\ref{simp}) and (\ref{actions}), the Q-function can be expanded and expressed by Taylor's theorem as
\begin{align}
Q^i(s,\bm{a}) 		=&  \frac{1}{|\mathcal{N}_i|}\sum\nolimits_{j} Q^i(s,a^i,a^j) \nonumber\\
=& \frac{1}{|\mathcal{N}_i|}\sum\nolimits_{j}  \bigg[  Q^i(s,a^i,\bar{a}^i)+\nabla_{\bar{a}^{i,j}}  Q^i(s,a^i,\bar{a}^i)\cdot \delta a^{i,j}\nonumber \\
&+\frac{1}{2}\delta a^{i,j}\cdot\nabla^2_{\bar{a}^{i,j}}Q^i(s,a^i,\tilde{a}^{i,j})\cdot\delta a^{i,j}\bigg] 		\nonumber\\
=& Q^i(s,a^i,\bar{a}^i) + \frac{1}{2{|\mathcal{N}_i|}} \sum\nolimits_{j} R_{s,a^i}^i (a^j) \nonumber\\
\approx & Q^i(s,a^i,\bar{a}^i),
\label{mfapprox}
\end{align}
where the second-order Taylor polynomial remainder $R^i_{s,a^i}(a^j)$ denotes $\delta a^{i,j} \cdot\nabla^2_{\bar{a}^{i,j}}Q^i(s,a^i,\tilde{a}^{i,j}) \cdot \delta a^{i,j}$ with $\tilde{a}^{i,j}=\bar{a}^i+\epsilon^{i,j}\delta a^{i,j}$ and $\epsilon^{i,j}\in[0,1]$.
Suppose the Q-function $Q^i(s,a^i,a^j)$ is $\Omega$-smooth, the remainder could be proved that it is bounded within the interval $[-2\Omega,2\Omega]$, and is dropped as a fluctuation term.

\section{Our Method}\label{method}
% In this section, we first propose a new protocol that efficiently learns local communication between neighboring agents via depthwise convolution.
In this section, we first propose a new protocol that efficiently learns local communication via depthwise convolution.
% Then, we present the enhanced mean-field approximation that uses a supervised policy tracking module to more accurately predict the real-time policies of neighboring agents.
Then, we present the enhanced mean-field approximation that uses PRN to rectify real-time agent interactions.
% Finally, we give the integrated algorithm based on the above-mentioned implementations.
Finally, we give the integrated algorithm.

\subsection{Depthwise Convolution-based Communication Protocol (DCCP)}
\begin{figure*}[tb]\centering
  \includegraphics[width=1.95\columnwidth]{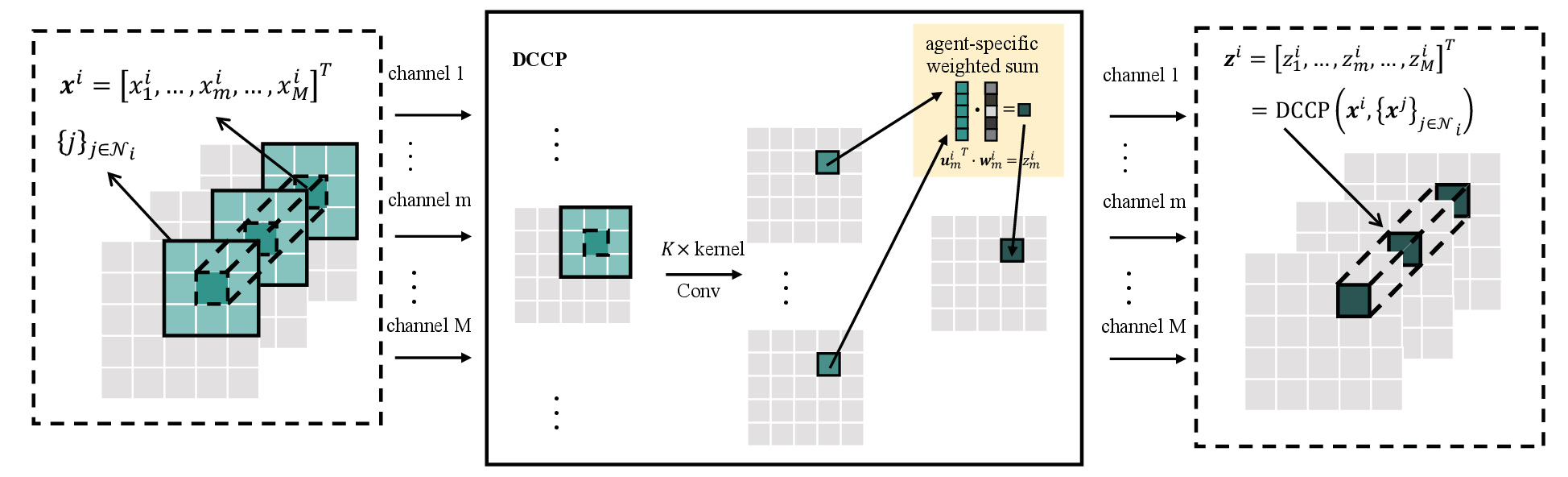}
  \caption{The illustration of DCCP. We use a different set of $K$ convolution kernels for each channel, and each kernel is shared for all agents. In channel $m$, the input of the $i$-th agent, $x_m^i$, is transformed to a $K$-dimensional vector $\bm{u}^i_m$ by convolution over the local area surrounding this agent. Then, the agent-specific weight vector $\bm{w}^i_m$ is used to calculate the weighted sum of $\bm{u}^i_m$ to obtain the final output $z_m^i$.}
  \label{fig1}
\end{figure*}

Due to the partial observability and the lack of accurate real-time interactions across agents, a communication protocol is vital to coordinate the behavior of each individual and to solve the task via sharing the observations and policies.
It is straightforward to broadcast essential messages across all agents, while the learning becomes intractable due to the exponential growth of agent interactions when the number of agents increases largely.
Moreover, this approach suffers from content redundancy and is unsustainable under bandwidth limitations. 
Instead, we consider learning a local communication protocol between neighboring agents.
Convolution is widely used in extracting local relations, and inspired by this, we introduce the spatial convolution into the local communication protocol.
In principle, the convolution can perform a form of system identification, conjecturing parameters of neighboring agents and coordinating the individual's behavior as a function of these parameters.

In MARL, the information to be shared usually has distinguishable channel-wise semantics.
For example, different channels of an observation vector may represent different attributions of the environment, or different channels in the output of a Q-network can express outcomes of different actions.
Standard convolution cannot preserve the channel-wise semantics since it filters and combines inputs from different channels into a new set of outputs in one step.
Therefore, we propose to use one-layer depthwise convolution~\cite{howard2017mobilenets} to filter inputs from the same channel and produce channel-wise outputs that are semantics-invariant with inputs.
In each channel, we share the parameters of multiple convolution kernels for all agents, aiming to extract and transfer the channel-wise common knowledge (embedded in the convolution kernels) across agents.
Then, each individual has its own agent-specific weights to calculate the weighted sum of outputs from these depthwise convolution kernels. 

Fig.~\ref{fig1} illustrates the proposed communication protocol via depthwise convolution.
In this paper, we assume that the neighboring relationship between agents keeps fixed during learning.
Let $\boldsymbol{x}^i=[x_1^i,x_2^i,...,x_M^i]^T$ denote the $M$-channel input vector of agent $i~(i=1,...,N)$, and let $\mathcal{N}_i$ denote the set of the $i$-th agent's neighbors.
For a given channel $m$, we use $K$ convolution kernels of size $n\times n$ to extract the local relations between agent $i$ and its neighbors $\{j\}_{j\in\mathcal{N}_i}$, which outputs a $K$-dimensional hidden vector $\bm{u}_m^i$ as $\bm{u}_m^i=\text{Conv}(x_m^i, \{x_m^j\}_{j\in\mathcal{N}_i})$.
We use a different set of $K$ convolution kernels for each channel, resulting in $M\cdot K$ convolution kernels in total.
Then, agent $i$ uses its own agent-specific weight vector $\bm{w}^i_m$ to calculate the weighted sum of outputs from the $K$ convolution kernels as $z_m^i={\bm{u}_m^i}^T\cdot\bm{w}_m^i$.
We gather the outputs of all $M$ channels to obtain the output vector of the Depthwise Convolution-based Communication Protocol (DCCP) as $\bm{z}^i=\text{DCCP}(\bm{x}^i, \{\bm{x}^j\}_{j\in\mathcal{N}_i})$.

Through the depthwise convolution channel by channel, the channel-wise semantics is well preserved and the output $\bm{z}^i$ of the communication protocol is semantics-invariant with the input $\bm{x}^i$.
% The convolution kernels are shared among all agents in order to modulate the common knowledge for coordinating behaviors between neighboring agents, while each individual maintains its agent-specific weights to combine these convolution kernels for promoting diversity of individual behaviors.
The convolution kernels are shared among all agents in order to modulate the common knowledge for behavior coordination, while each individual maintains its agent-specific weights to combine the convolution kernels for promoting diversity of individual behaviors.

\subsection{Enhanced Mean-Field Approximation}\label{sec42}
In MARL, the learning of an agent's optimal policy depends on the dynamics of the others that are a part of the environment.
In the paper, we use the mean-field approximation~\cite{mean} to model the behaviors of other agents for each individual.
It approximately treats the interactions within agents as the interaction between an individual and a virtual agent averaged by other agents, which transmits messages across agents with a reduced scale of agent interactions.

In mean-field Q-learning (MF-Q), the $i$-th agent's Q-function of the global joint action is first factorized using only pairwise interactions between neighboring agents as
\begin{equation}
\label{mq1}
Q^i(\bm{s},\bm{a})=\frac{1}{|\mathcal{N}_i|}\sum\nolimits_{j\in \mathcal{N}_i}Q^i(\bm{s},a^i,a^j),
\end{equation}
where $\bm{s}$ is the global state, $\bm{a}$ is the joint action of agents, and $|\cdot|$ denotes the cardinality of the set.
By the Taylor's theorem, the Q-function can be expanded and approximated as
\begin{equation}
\label{mq2}
Q^i(\bm{s}, \bm{a}) \approx Q^i\left(\bm{s}, a^i, \frac{1}{|\mathcal{N}_i|}\sum\nolimits_{j\in\mathcal{N}_i}a^j\right).
\end{equation}
Since an individual cannot know real-time policies of its neighboring agents, MF-Q uses previous actions of the neighbors to estimate the current action $\tilde{a}_t^i$ as
\begin{equation}\label{mf11}
\tilde{a}_t^i = \mu^i_t\left(\bm{s}_t, \frac{1}{|\mathcal{N}_i|}\sum\nolimits_{j\in \mathcal{N}_i}a_{t-1}^j\right),
\end{equation}
where $\mu^i$ is the policy function derived from $Q^i$.
In place of the real action $a_t^j$, the estimated one $\tilde{a}_t^j$ is used to make the real-time decision $a_t^i$ as
\begin{equation}\label{mf22}
a_t^i = \mu^i_t\left(\bm{s}_t, \frac{1}{|\mathcal{N}_i|}\sum\nolimits_{j \in \mathcal{N}_i}\tilde{a}^j_{t}\right).
\end{equation}
However, the estimated action $\tilde{a}_t$ has no guaranteed similarity with the real one $a_t$ for two reasons: 1) the same policy function $\mu$ is used in both action estimation and real decision making; 2) the estimated action $\tilde{a}_t$ is predicted from previous real action $a_{t-1}$ that may have no explicit correlation with the current real action $a_t$.
Moreover, MF-Q assumes that each agent has access to the global state of the system, which might be problematic in multi-agent settings where each agent can only observe a local part of the shared environment.

\begin{figure}[tb]\centering
	\includegraphics[width=0.75\columnwidth]{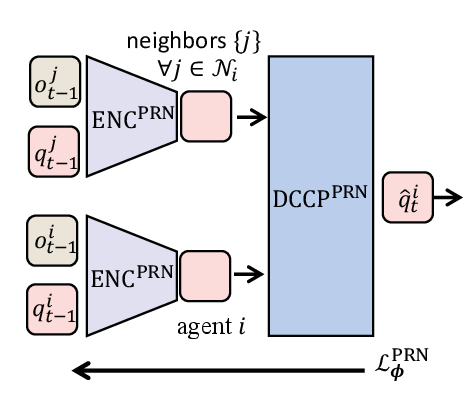}
	\caption{The illustration of the policy rectification network (PRN).}
	\label{fig2}
\end{figure}

In order to more accurately coordinate behaviors of neighboring agents, we enhance the mean-field approximation by a supervised PRN that predicts real-time actions from previous information to rectify real-time agent interactions.
Since the policy is derived from the Q-values, we estimate the agent's real-time Q-values $\hat{\bm{q}}_t^i$ from previous partial observations $\bm{o}_{t-1}$ and Q-values $\bm{q}_{t-1}$ of itself and its neighboring agents as
\begin{equation}
\hat{\bm{q}}_t^i = f_{\bm{\phi}}^{\text{PRN}}\left((\bm{o}_{t-1}^i, \bm{q}_{t-1}^i), \{(\bm{o}_{t-1}^j, \bm{q}_{t-1}^j)\}_{j\in\mathcal{N}_i}\right),
\end{equation}
where $f_{\bm{\phi}}^{\text{PRN}}$ is the PRN function parameterized by weights $\bm{\phi}$.
It is trained in a supervised regression manner to make the estimated Q-values $\hat{\bm{q}}_t^i$ more accurate about the ground truth $\bm{q}_t^i$, and the loss function is formalized as 
\begin{equation}
\mathcal{L}^{\text{PRN}}_{\bm{\phi}} = \sum\nolimits_{i\in N} ||\bm{q}_t^i-\hat{\bm{q}}_t^i||^2.
\end{equation}

Fig.~\ref{fig2} illustrates the network structure of PRN in detail.
First, we use an encoder that is shared across agents to embed the input $(\bm{o}^i_{t-1},\bm{q}_{t-1}^i)$ into a latent vector $\bm{v}^i$ with the same dimension as the action space as $\bm{v}^i = \text{ENC}^{\text{PRN}}(\bm{o}^i_{t-1},\bm{q}_{t-1}^i)$. 
Then, we feed the latent vectors of agent $i$ and its neighbors into our DCCP for information sharing and behavior coordination, and obtain the predicted real-time Q-values $\hat{\bm{q}}_t^i$ as $\hat{\bm{q}}_t^i = \text{DCCP}^{\text{PRN}}(\bm{v}^i, \{\bm{v}^j\}_{j\in\mathcal{N}_i})$.  
In short, the PRN function can be described as
\begin{equation}
\label{ept}
\hat{\bm{q}}_t^i \!=\! \text{DCCP}^{\text{PRN}}\!\!\left(\!\text{ENC}^{\text{PRN}}\!\!\left(\!(\bm{o}_{t-1}^i,\bm{q}_{t-1}^i),\! \{(\bm{o}_{t-1}^j,\bm{q}_{t-1}^j)\}_{j\in\mathcal{N}_i}\!\right)\!\right)\!.
\end{equation}

\subsection{The Integrated Algorithm}\label{sec43}
\begin{figure}[tb]\centering
	\includegraphics[width=0.99\columnwidth]{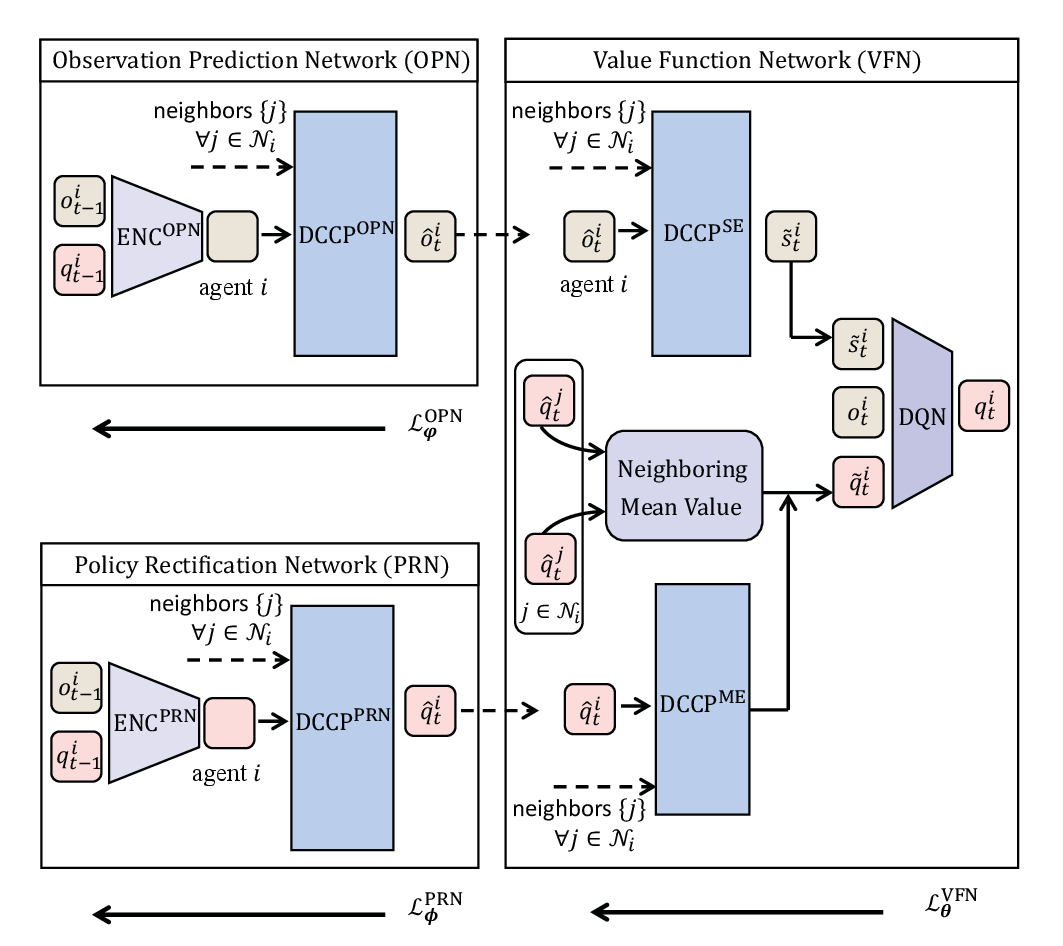}
	\caption{The network architecture of our method. }
	\label{fig3}
\end{figure}

With the above, Fig.~\ref{fig3} depicts the architecture of our method that consists of three parts: the observation prediction network (OPN) with DCCP, the PRN with enhanced mean-field approximation, and the value function network (VFN).

The first module, OPN, employs DCCP to share observations across neighboring agents, which addresses the partial observability issue with low computational cost.
In real-world applications, the communication of real-time information may lead to a certain time lag due to bandwidth limitations~\citep{tarmac}.
Analogous to PRN, we predict real-time observations from previous information to more accurately coordinate behaviors of neighboring agents.
In detail, we use the previous observations $\bm{o}_{t-1}$ and the previous Q-values $\bm{q}_{t-1}$ of an agent and its neighbors to predict the agent's real-time observation $\hat{\bm{o}}_t^i$ as
\begin{equation}
\hat{\bm{o}}_t^i = f_{\bm{\varphi}}^{\text{OPN}}\left((\bm{o}_{t-1}^i, \bm{q}_{t-1}^i), \{(\bm{o}_{t-1}^j, \bm{q}_{t-1}^j)\}_{j\in\mathcal{N}_i}\right),
\end{equation}
where $f_{\bm{\varphi}}^{\text{OPN}}$ denotes the OPN function parameterized by weights $\bm{\varphi}$. 
This prediction problem is analogous to learning the state transition model since the Q-values can be considered as the action information equivalently, and is trained in a supervised regression manner using the current observation $\bm{o}_t^i$ as the ground truth as
\begin{equation}
\mathcal{L}^{\text{OPN}}_{\bm{\varphi}} = \sum\nolimits_{i\in N}||\bm{o}_t^i-\hat{\bm{o}}_t^i||^2.
\end{equation}
We use another encoder that is also shared across agents to embed the agent's input $(\bm{o}_{t-1}^i, \bm{q}_{t-1}^i)$ into a latent vector with the same dimension as the observation space. 
Then, we feed the latent vectors of an agent and its neighbors into our DCCP to obtain the predicted real-time observation as 
\begin{equation}
\label{eop}
\hat{\bm{o}}_t^i \!=\!\text{DCCP}^{\text{OPN}}\!\!\left(\!\text{ENC}^{\text{OPN}}\!\!\left(\!(\bm{o}_{t-1}^i,\bm{q}_{t-1}^i),\! \{(\bm{o}_{t-1}^j,\bm{q}_{t-1}^j)\}_{j\in\mathcal{N}_i}
\!\right)\!\right)\!.
\end{equation}

The PRN has been described in the last subsection, and finally, we present the VFN. It evaluates the Q-values $\bm{q}_t^i$ from the predicted real-time observations $\bm{\hat{o}}_t$ and the rectified real-time agent interactions $\bm{\hat{q}}_t$ of an agent and its neighbors, and the agent's real-time observation $\bm{o}_t^i$ as 
\begin{equation}
{\bm{q}}_t^i = f_{\bm{\theta}}^{\text{VFN}}\left(\bm{o}_t^i, (\hat{\bm{o}}_{t}^i, \hat{\bm{q}}_{t}^i), \{(\hat{\bm{o}}_{t}^j, \hat{\bm{q}}_{t}^j)\}_{j\in\mathcal{N}_i}\right),
\end{equation}
where $f_{\bm{\theta}}^{\text{VFN}}$ denotes the VFN function parameterized by weights $\bm{\theta}$, and $\bm{\theta}$ are shared across agents except for the agent-specific weights in the two DCCPs.

More concretely, VFN contains three parts: the observation sharing for global state estimation, the policy sharing for compensating the mean-field approximation bias, and the deep Q-network (DQN). 
First, we feed the predicted real-time observations $\hat{\bm{o}}_t$ of an agent and its neighbors for information sharing and coordination, and obtain the global state estimation (SE) $\tilde{\bm{s}}_t^i$ as 
\begin{equation}
\label{eoc}
\tilde{\bm{s}}_t^i = \text{DCCP}^{\text{SE}}\left(\hat{\bm{o}}_t^i, \{\hat{\bm{o}}_t^j\}_{j\in\mathcal{N}_i}\right).
\end{equation}

% Second, to simplify agent interactions, MF-Q uses the Taylor's expansion to approximate the Q-function in~(\ref{mq2}) and drops out the second-order remainders.
Second, to simplify agent interactions, MF-Q expands the Q-function in~(\ref{mq2}) using Taylor's theorem and drops out the second-order remainders.
We attempt to exploit DCCP to compensate for this approximation bias by sharing the predicted real-time Q-values $\hat{\bm{q}}_t$ across neighboring agents, and obtain a more accurate mean-field estimate (ME) $\tilde{\bm{q}}^i_t$ as
\begin{equation}
\tilde{\bm{q}}^i_t = \frac{1}{|\mathcal{N}_i|}\sum\nolimits_{j\in\mathcal{N}_i}\hat{\bm{q}}_t^j + \text{DCCP}^{\text{ME}}\left(\hat{\bm{q}}_t^i, \{\hat{\bm{q}}_t^i\}_{j\in\mathcal{N}_i}\right),
\end{equation}
% The enhanced estimate consists of the original mean-value of neighboring agents $\frac{1}{|\mathcal{N}_i|}\sum_{j\in\mathcal{N}_i}\hat{\bm{q}}_t^j$ plus a DCCP-based compensation term.
which consists of the rectified mean-value of neighboring agents plus a DCCP-based compensation term.
Since this DCCP is trained using the DQN loss directly, it has the potential to implicitly compensate for the second-order remainders in an end-to-end manner.

% Third, the DQN is composed of a fully-connected MLP with shared parameters for all agents, and takes the concatenation of the real-time observation $\bm{o}_t^i$, the estimated global state $\tilde{\bm{s}}_t^i$, and the enhanced mean-field estimate $\tilde{\bm{q}}_t^i$ as input as
Third, the DQN takes the concatenation of the real-time observation $\bm{o}_t^i$, the estimated global state $\tilde{\bm{s}}_t^i$, and the enhanced mean-field estimate $\tilde{\bm{q}}_t^i$ as input as
\begin{equation}
\bm{s}_t^i = \left[\bm{o}_t^i, \tilde{\bm{s}}_t^i,\tilde{\bm{q}}_t^i \right].
\end{equation}		
The output of the DQN is denoted as $\bm{q}_t^i = Q(\cdot|\bm{s}_t^i)$, and the loss function is formalized as the Bellman residual as
\begin{equation}
\mathcal{L}_{\bm{\theta}}^{\text{VFN}} = \sum\nolimits_{i\in N}\left(y_t^i - Q\left( {a}_t^i\right | \bm{s}_t^i)\right)^2,
\end{equation}
in which the bootstrapped target $y_t^i$ is calculated as 
\begin{equation}
\label{y}
y^i_t = \left\{
\begin{aligned}
&r^i_t, & \text{if~terminate}, \\
&r^i_t+\gamma\max\nolimits_{{a}'} Q^{\text{target}}({a}' | \bm{s}_{t+1}^i)  , & \text{otherwise},
\end{aligned}
\right.
\end{equation}
where $Q^{\text{target}}$ is the target network with parameters copied from some previous version of the DQN.

Together, the loss function of our method is aggregated as 
\begin{equation}
\label{loss}
\mathcal{L}\left(\bm{\theta},\bm{\phi},\bm{\varphi}\right) 
= \mathcal{L}^{\text{VFN}}_{\bm{\theta}}+\lambda_{1}\mathcal{L}^{\text{PRN}}_{\bm{\phi}}+\lambda_{2}\mathcal{L}^{\text{OPN}}_{\bm{\varphi}},
\end{equation}
where $\lambda_1$ and $\lambda_2$ are coefficients that balance the influence of three modules' loss functions.
\footnote{The supervised prediction of OPN and PRN involves online learning, which can be stabilized by the experience replay mechanism, analogous to the training of deep Q-network~\citep{dqn}.
Moreover, we may increase the update frequency of OPN and PRN to stabilize the supervised prediction modules before focusing on training the RL module VFN.}
Correspondingly, the integrated algorithm is summarized as shown in Algorithm 1.
The network parameters $\bm{\theta}$, $\bm{\phi}$, and $\bm{\varphi}$ are shared across agents except for the agent-specific weights $\bm{w}^i$.
Specifically, the agent-specific weights $\bm{w}^i$ are updated as
\begin{equation}
\begin{aligned}
\bm{w}^i \leftarrow & \bm{w}^i - \alpha\nabla_{\bm{w}^i} \bigg[\left( y^i_{\tau} - Q(\bm{s}^i_{\tau},a^i_{\tau}) \right)^2   \\
& + \lambda_1 ||\bm{q}_{\tau}^i-\hat{\bm{q}}_{\tau}^i ||^2  + \lambda_2 ||\bm{o}_{\tau}^i-\hat{\bm{o}}_{\tau}^i ||^2\bigg], ~~\forall i \in N.
\end{aligned}
\end{equation}

\begin{algorithm}[tb]
\caption{DCCP-based MARL with enhanced mean-field approximation}
\label{alg1}
\begin{algorithmic}[1]
	\STATE Initialize exploration ratio $\epsilon$, learning rate $\alpha$, \newline and coefficients $\lambda_1, \lambda_2$ \\ 
	\STATE Define a transition $e_t^i~(i=1,...,N)$ in replay buffer $\mathcal{B}$ as 
	\begin{align}\nonumber 
	e_t^i = \{\bm{o}_{t-1}^i,\bm{q}_{t-1}^i,\bm{o}_t^i,\bm{a}_t^i,r_t^i,\bm{q}_t^i,\bm{o}_{t+1}^i\} 
	\end{align}  \\
	\STATE Randomly initialize the parameters $\bm{\phi}$, $\bm{\varphi}$, $\bm{\theta}$ \\
	\WHILE{not converge}
		\STATE Initialize $\bm{q}_0^i$ and $\bm{o}_0^i$ for each agent $i$ \\
        \FOR {$t=1,...,T$ (terminal)}
    		\FOR {each agent $i = 1,...,N$}
    			\STATE	With probability $\epsilon$ select a random action $a_t^i$, \newline otherwise $a_t^i=\arg\max_a \bm{q}_t^i$ \\
    		\ENDFOR 
    		\STATE	Take joint action $[a_t^1,...,a_t^N]$, obtain $r_t^i$ and $\bm{o}_{t+1}^i$
    		\STATE	Store transitions $e_t^i$ into $\mathcal{B}$ \\
    	\ENDFOR
		\STATE	Sample a minibatch of transitions $e_{\tau}^i $ from $\mathcal{B}$ \\
		\STATE	Calculate $\hat{\bm{o}}_{\tau}^i$, $\hat{\bm{q}}_{\tau}^i$, and $y_{\tau}^i$ using~(\ref{eop}), (\ref{ept}), and (\ref{y}) \\
		% \STATE	Set $y_{\tau}^i$ according to~(\ref{y}) \\
		\STATE	Perform a gradient descent step as 
			\begin{align}
			&\bm{\theta}\leftarrow\bm{\theta} - \alpha \sum\nolimits_{i\in N} \nabla_{\bm{\theta}}\left( y^i_{\tau} - Q(\bm{s}^i_{\tau},a^i_{\tau}) \right)^2 \nonumber\\
			&\bm{\phi}\leftarrow\bm{\phi} - \alpha\lambda_1 \sum\nolimits_{i\in N} \nabla_{\bm{\phi}}|| \bm{q}_{\tau}^i-\hat{\bm{q}}_{\tau}^i ||^2 \nonumber\\		
			&\bm{\varphi}\leftarrow\bm{\varphi} - \alpha\lambda_2 \sum\nolimits_{i\in N} \nabla_{\bm{\varphi}}|| \bm{o}_{\tau}^i-\hat{\bm{o}}_{\tau}^i ||^2 \nonumber	
			% &\textcolor{red}{\bm{w}^i\leftarrow\bm{w}^i - \alpha\nabla_{\bm{w}^i} \bigg[\left( y^i_{\tau} - Q(\bm{s}^i_{\tau},a^i_{\tau}) \right)^2}  \nonumber \\
			% & \textcolor{red}{ \qquad + \lambda_1 ||\bm{q}_{\tau}^i-\hat{\bm{q}}_{\tau}^i ||^2 + \lambda_2 ||\bm{o}_{\tau}^i-\hat{\bm{o}}_{\tau}^i ||^2\bigg], \forall i \in N} \nonumber
			\end{align}\\
		\STATE Update $Q^{\text{target}}\leftarrow Q$ every $\eta$ steps\\
	\ENDWHILE
\end{algorithmic}
\end{algorithm}

\section{Experiments}\label{experiments}
We conduct experiments applying the proposed method to a mixed-cooperative ATSC task on the SUMO platform and to a fully-cooperative SMAC task, to imply the method's ability in general cooperative circumstances to achieve efficient coordination to achieve high returns. 
We compare our method to the following baselines: the CTDE-based QMIX~\cite{rashid2018qmix, rashid2020weighted} and the global communication learning-based NDQ~\cite{ndq} for the fully-cooperative SMAC; the global communication learning-based TarMAC~\cite{tarmac} and the local communication learning-based IC3Net~\cite{ic3net} for both tasks.
\footnote{
The value function factorization methods, e.g., QMIX and NDQ, use a mixture network to predict the global value function and are only feasible for fully-cooperative tasks. 
Hence, we do not evaluate them in the mixed-cooperative ATSC environments where each agent has its own objective of controlling the local traffic situation.}
Then, we perform an ablation study to verify the respective effectiveness of the two components in our method.
IQL~\cite{iql} is set as the baseline approach that removes the two components from our method, and DCCP is a variant of our method that only removes the enhanced mean-field approximation.
The effect of DCCP is demonstrated by comparing DCCP with IQL, and the effect of the enhanced mean-field approximation is demonstrated by comparing our method with DCCP.
Since the mean-field approximation is partly enhanced by DCCP, we do not consider the variant of only removing the DCCP component.
All results are averaged over $10$ seeds. 
The shaded area represents the 95\% confidence interval for evaluation curves, and the standard errors are presented for numerical results.

\subsection{Adaptive Traffic Signal Control (ATSC)}
\begin{figure}[tb]
	\centering
	\subfigure[A traffic gird of 25 intersections with 4 example flows.]{
		\includegraphics[width=0.9\linewidth]{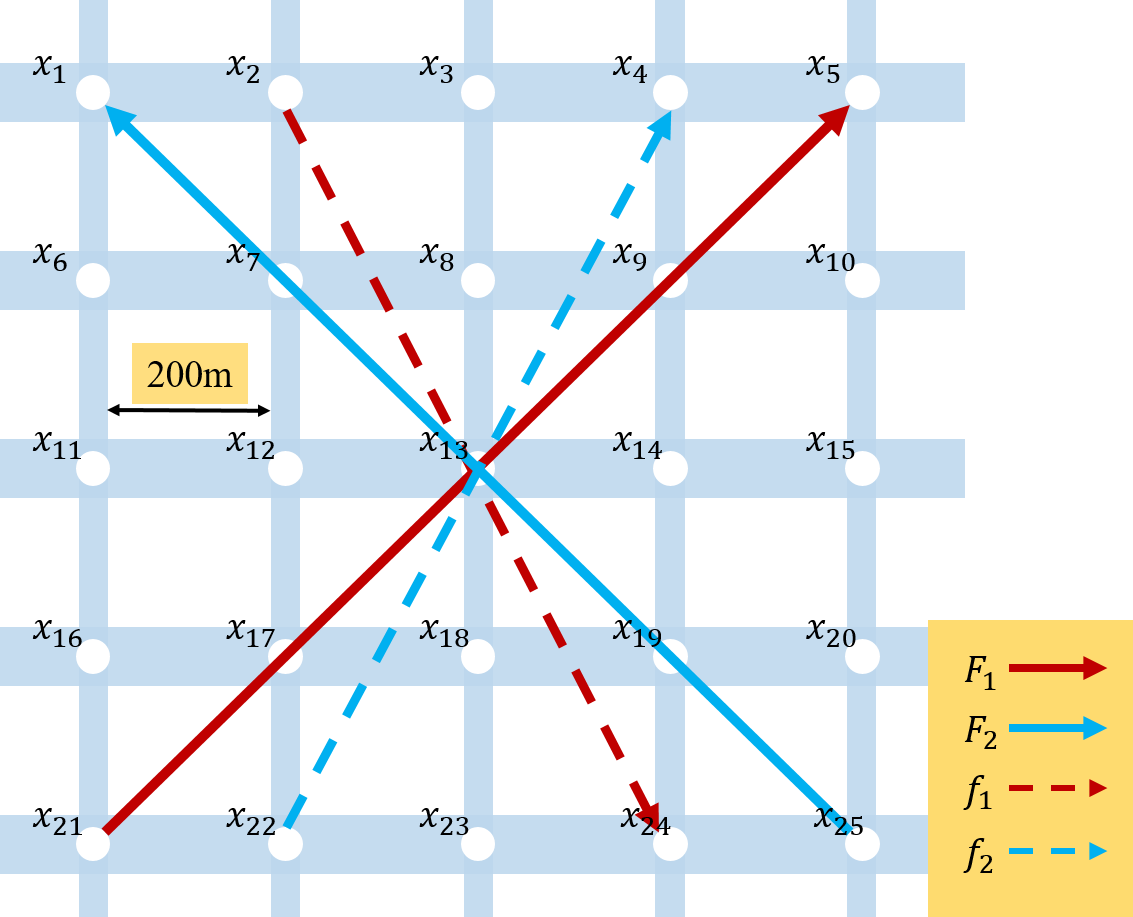}
		\label{setting_a}}
	\subfigure[Intersection.]{
		\includegraphics[width=0.3\linewidth]{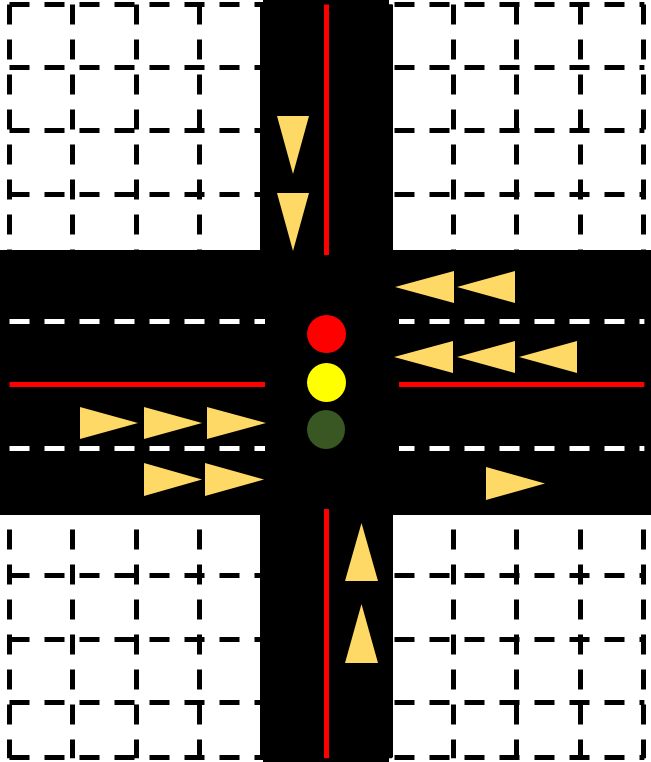}
		\label{setting_b}}
	\centering
	\subfigure[Possible actions.]{
		\includegraphics[width=0.55\linewidth]{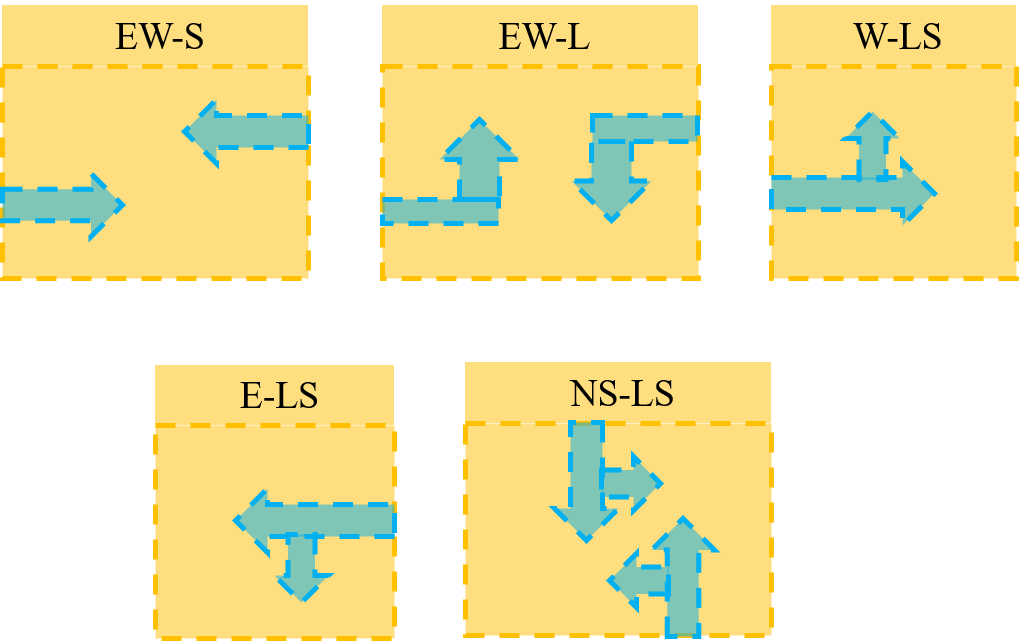}
		\label{setting_c}}
		\caption{Simulation environment of ATSC.}
		\label{setting}
\end{figure}

We conduct ATSC in a  $5\!\times\! 5$ synthetic grid map on the standard traffic simulation platform SUMO, where each agent (traffic intersection) observes a local part of the shared environment and controls phases of its traffic signals.
The objective of ATSC is to eliminate the traffic congestion in each intersection to generate low congestion of the whole traffic system, i.e., minimizing the queue length and time delay.
The mixed cooperation occurs as each agent takes the traffic situations in its local intersection and in the whole traffic network as the objective.
Fig.~\ref{setting_a} shows a synthetic traffic network that consists of multiple homogeneous intersections in a $5\times 5$ grid map.
Let E, N, W, and S denote east, north, west, and south, respectively.
Each intersection consists of two E-W two-lane streets and two N-S one-lane avenues, as shown in Fig.~\ref{setting_b}.
Each intersection has five available phases (corresponding to the actions in RL settings) that are EW-S, EW-L, W-LS, E-LS, and NS-LS, as shown in Fig.~\ref{setting_c}. 

\begin{figure}[tb]\centering
	\includegraphics[width=0.8\columnwidth]{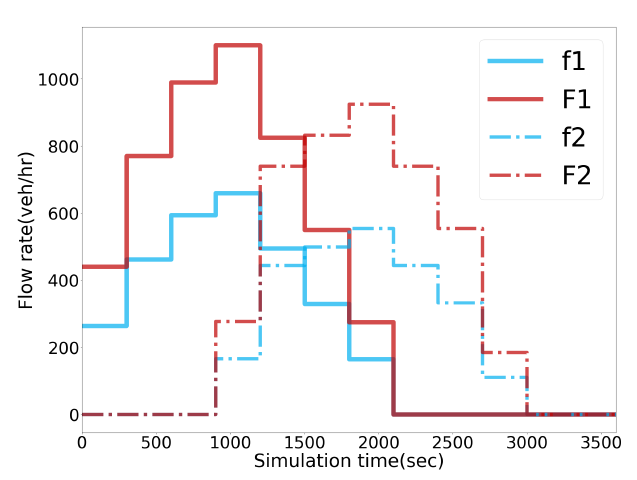}
	\caption{An example of the time-varying function of traffic flows in ATSC.}
	\label{trafficflow}
\end{figure}

Following the same experiment setting in~\citep{ma2c,nmarl}, we use the traffic flow to describe the vehicles in the simulation. The traffic flow corresponds to generating vehicles arriving at the traffic network with an origin and a destination, i.e., an origin-destination (O-D) pair.
An O-D pair is denoted as $x_j$-$x_k$, where the origin $x_j$ and the destination $x_k$ denote two different intersections in the traffic network as shown in Fig.~\ref{setting_a}.
In the simulation, the peak-hour traffic dynamics is constituted by four time-varying traffic flows: $F_1$, $F_2$, $f_1$, and $f_2$, each of which generates vehicles according to three different O-D pairs.
We show the flows $F_1$ and $f_1$ and their O-D pairs in Fig.~\ref{setting_a}.
$F_1$ consists of three O-D pairs across the main streets: $x_1$-$x_{25}$, $x_{11}$-$x_{15}$, and $x_{21}$-$x_5$.
$f_1$ consists of three O-D pairs across the side avenues: $x_2$-$x_{24}$, $x_3$-$x_{23}$, and $x_4$-$x_{22}$.
Similarly, the flows $F_2$ and $f_2$ generate vehicles according to O-D pairs that are opposite to those of $F_1$ and $f_1$, respectively.
In the simulation, the vehicles are generated from the four traffic flows with time-varying flow rates.
Fig.~\ref{trafficflow} presents the flow rate function in our simulation of the four traffic flows.

The local observation of the $i$-th agent is defined as 
\begin{equation}
o_t^i = \left\{  \left(\text{time\_delay}_t\left[ l \right] \text{,} \text{wave}_t\left[ l \right] \right) \right\}_{l\in L_i},
\end{equation}
where $l$ is an incoming lane of the $i$-th intersection, and $L_i$ is the set of all incoming lanes of intersection $i$. 
$ \text{time\_delay}_t\left[ l \right] \left(\text{s}\right)$ measures the cumulative delayed time of the first vehicle in lane $l$ at step $t$, and  $\text{wave}_t\left[ l \right]\left(\text{veh}\right)$ measures the total number of approaching vehicles along lane $l$ within $50$ meters to intersection $i$ at step $t$.
The action of the $i$-th agent is represented as a one-hot vector that denotes selecting one traffic phase from the five available phases in Fig.~\ref{setting_c} as
\begin{equation}
a_t^i\in\{\text{EW-S, EW-L, W-LS, E-LS, NS-LS}\}. 
\end{equation}
Specifically, when the next action is different from the last one, an all-yellow phase will appear and last for $2$ seconds to ensure a safe switch between different phases.
The reward of the $i$-th agent is defined as the weighted sum of the queue length and time delay of the vehicles at the corresponding intersection as
\begin{equation}
r_t^i = - \!\sum\nolimits_{l\in L_i} \!\!\left( \text{queue\_len}_{t+1}\left[ l \right] \!+\! w \!\cdot\! \text{time\_delay}_{t+1}[l]  \right)\!,
\end{equation}
where $w$ is the trade-off coefficient that is set as $0.2$,
$\text{queue\_len}_{t+1}[l]$ is the measured number of vehicles in the waiting queue, and the $\text{time\_delay}_{t+1}[l]$ is the cumulative delay time (in seconds) of the first vehicle along each incoming lane $l$ at the next time step.

\begin{table}[tb]
\setlength{\tabcolsep}{1.5mm}
\renewcommand\arraystretch{1.1}
	\centering
	\caption{Training hyperparameters of our method in ATSC.}
	\begin{tabular}{c|cccccc}
		\toprule
		Task		& replay buffer size    &  batch size    & $\lambda_1$   & $\lambda_2$   & $\alpha$ & optimizer \\
		\midrule	
		ATSC	    & $4000$        & $240$        & $0.5$        & $0.5$ & $1$e-$6$  & RMSProp  \\
		\bottomrule
	\end{tabular}
	\label{atsc_hyper}
\end{table}

We set the episode length to $720$ where each learning step takes $5$s in the simulator.
After training for $1$M steps, we evaluate the tested algorithms for one episode.
The performance metrics are the queue length and time delay at the traffic intersections in evaluation.
For each metric, we record the average value over all steps of the evaluation episode and the final value at the end of the evaluation.
The kernel size in our DCCP is fixed as $3\times 3$, and each agent can only access the information of its neighbors within the $3\times 3$ area.
Table~\ref{atsc_hyper} and Table~\ref{atsc_net} present training hyperparameters and the network architecture, respectively.

\begin{figure}[tb]
	\includegraphics[width=0.9\columnwidth]{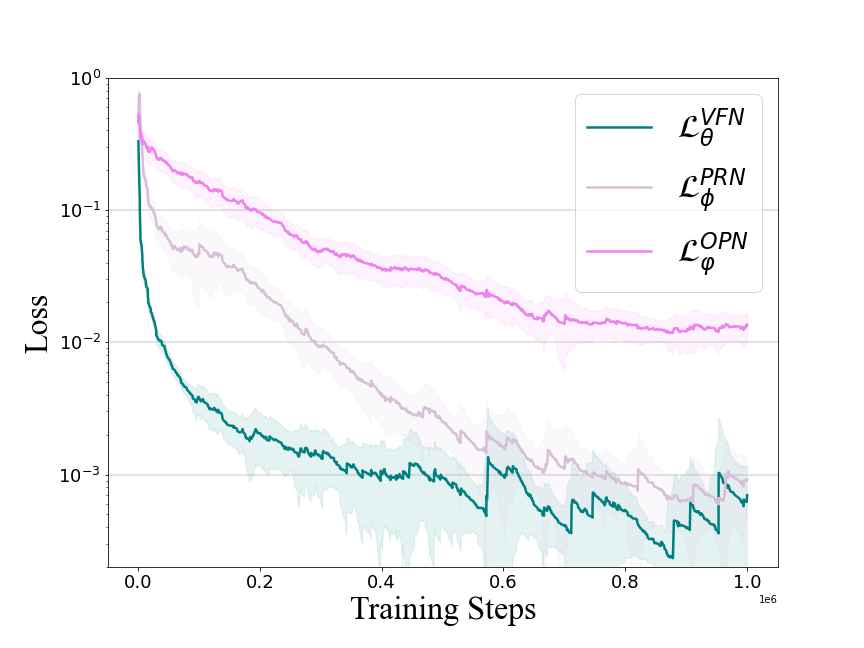}
	\caption{The training losses of the three modules of our method in ATSC.}
	\label{fig:atsc_loss}
\end{figure}

\begin{table}[t]
\setlength{\tabcolsep}{1.5mm}\renewcommand\arraystretch{1.1}
	\centering
	\caption{The network architecture of our method in ATSC.}
	\begin{tabular}{c|ccc|cc}
		\toprule
		\multirow{2}{*}{Modules}     & \multicolumn{3}{c|}{Encoder - layer size}   &\multicolumn{2}{c}{DCCP} \\
		& input    &  hidden  & output   & \# kernels    & kernel size \\
		\midrule	
		OPN	    &      $17$ ($12+5$)         & $256$			& $12$			&$10$           & $3\times3$       \\
		PRN   		&      $17$ ($12+5$)        & $256$			& $5$			&$10$           & $3\times3$       \\
		VFN		    	&      $29$ ($12+12+5$)         & $128$			& $5$			&$10$           & $3\times3$       \\
		
		\bottomrule
	\end{tabular}
	\label{atsc_net}
\end{table}

\begin{figure*}[tb]
	\centering
	\subfigure[received return]{\includegraphics[width=0.32\textwidth]{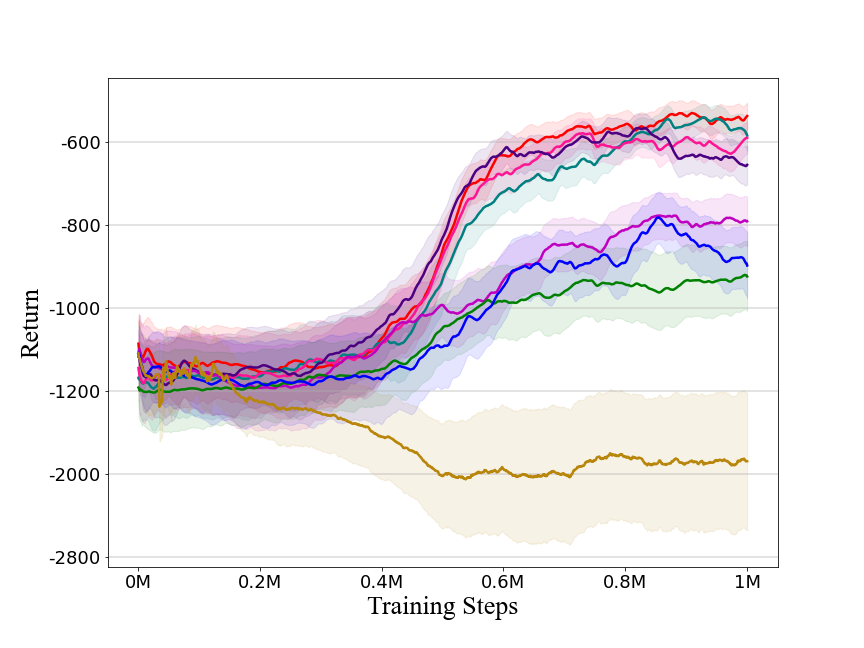}}	
	\subfigure[queue length]{\includegraphics[width=0.32\textwidth]{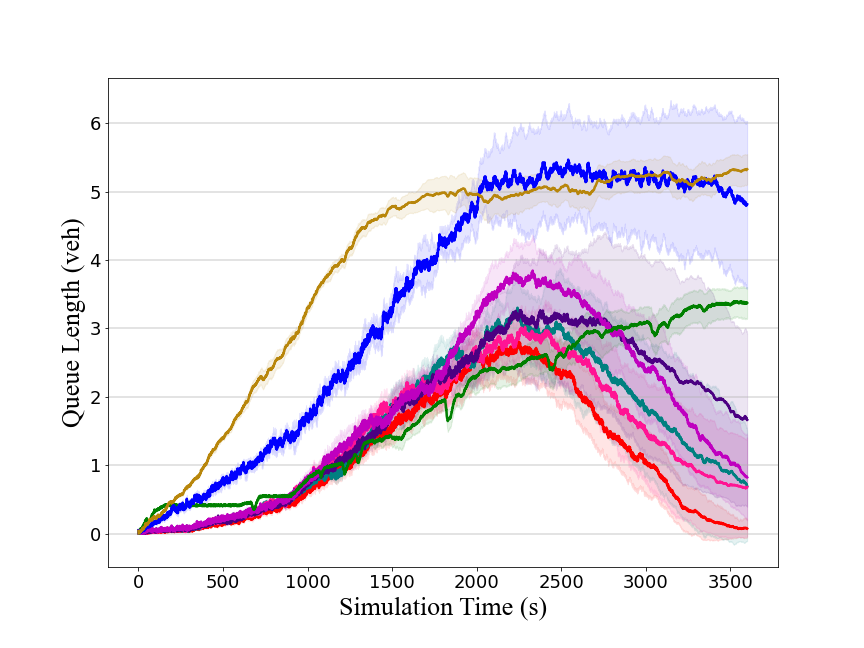}}
	\subfigure[time delay]{\includegraphics[width=0.32\textwidth]{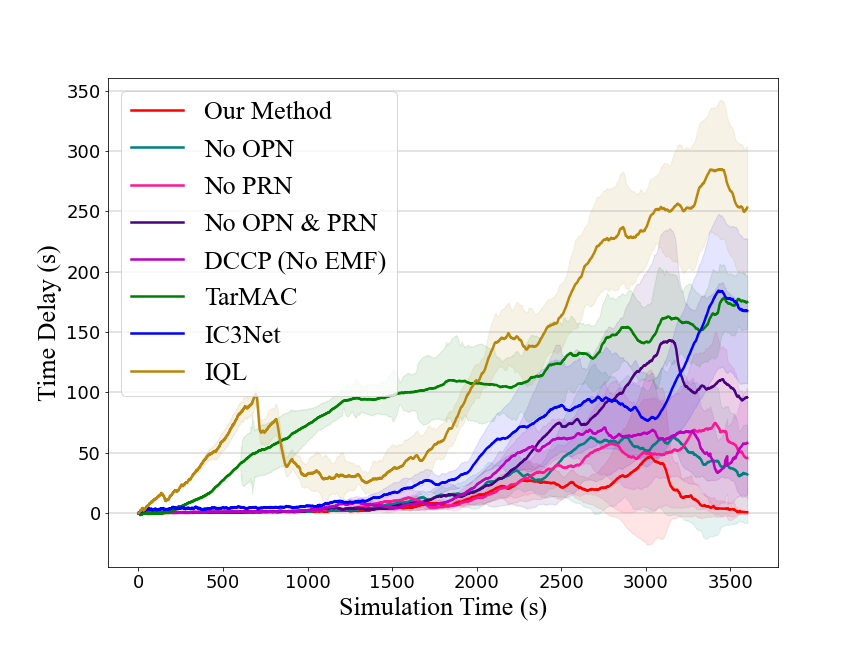}}	
	\caption{Evaluation results of the received return, queue length, and time delay in ATSC.}
	\label{fig:atsc_eva}
\end{figure*}

\begin{table*}[tb]
\renewcommand\arraystretch{1.1}
	\centering
	\caption{Numerical results of the average queue length, final queue length, average time delay, and final time delay in ATSC.}
	\begin{tabular}{l|rr|rr}
		\toprule
		Methods		& avg\_queue\_len (veh)       & final\_queue\_len (veh)    & avg\_time\_delay (s)      & final\_time\_delay (s) \\
		\midrule	
		IC3Net	    & $3.39\pm 0.27$        & $4.84\pm 0.76$        & $66.42\pm 9.26$       & $218.91\pm 25.43$ \\
		TarMAC      & $1.80\pm 0.06$   & $3.37\pm 0.11$        & $97.06\pm 3.23$       & $174.96\pm 11.45$ \\
		IQL		    & $3.91\pm 0.07$		& $5.32\pm 0.11$        & $118.35\pm 7.01$      & $253.22\pm 26.12$ \\
		\midrule
		DCCP (No EMF)	    & $1.76\pm 0.10$        & $0.87\pm 0.32$        & $27.30\pm 4.51$       & $58.14\pm 23.42$ \\
		No OPN & $1.50\pm 0.14$ & $0.69\pm 0.40$ & $22.58\pm 6.21$ &  $32.12\pm 13.55$ \\
		No PRN & $1.40\pm 0.11$ & $0.68\pm 0.36$ & $22.17\pm 5.76$ &  $45.82\pm 15.57$ \\
		No OPN \& PRN & $1.73\pm 0.19$ & $1.66\pm 0.58$ & $40.03\pm 9.29$ &  $95.81\pm 19.96$ \\
		Our Method	& $\bm{1.16\pm 0.08}$    & $\bm{0.00\pm 0.00}$   & $\bm{8.67\pm 1.72}$   & $\bm{0.03\pm 0.03}$ \\
		\bottomrule
	\end{tabular}
\label{tab:atsc_eva}
\end{table*}

First, we simply record the communication overhead.
IQL incurs no communication, and TarMAC takes $24*25$ rounds of communication as each agent needs to receive the information from the other $24$ agents.
In contrast, IC3Net and our method only cost $8*25$ rounds of communication as each agent only needs to receive the information from its 8 neighbors within the convolution kernel.
It can be observed that local communication saves communication resources compared to the global protocol, especially when the scale of the neighborhood is far smaller than the total scale.

Fig.~\ref{fig:atsc_loss} shows the training loss curves of the OPN, PRN, and VFN modules in our method.
It can be observed that the three modules are stably trained as the losses generally decrease during training.
Fig.~\ref{fig:atsc_eva} presents the evaluation curves of the received return, queue length, and time delay at the intersections, and Table~\ref{tab:atsc_eva} gives the numerical results in terms of the average and final values in evaluation.
Overall, our method performs the best and solves the ATSC task with its efficient coordination.
Using our method, the traffic congestion quickly decreases after peak flows, and is almost eliminated at the end as both the final queue length and time delay are approximately zero.
IC3Net obtains unsatisfactory performance as the traffic congestion quickly increases after peak flows and remains at a high level for a long time.
Since IC3Net simply averages the messages from neighboring agents, it cannot differentiate valuable information that helps cooperative decision making. 
TarMAC performs slightly better than IC3Net, which is supposed to benefit from allowing each agent to actively select which agents to address messages to.

Next, we conduct a comprehensive ablation study. 
IQL performs the worst due to the lack of communication across agents.
The distinct superiority of DCCP over IQL verifies the effectiveness of our communication protocol.
Moreover, DCCP mostly obtains better performance than IC3Net and TarMAC, which demonstrates the advantage of our communication protocol over other communication learning approaches.
At last, the performance improvement of our method over DCCP shows that the enhanced mean-field approximation can further facilitate multi-agent coordination.
Further, we perform the ablation study to investigate the effects of the OPN and PRN modules on our method, corresponding to the variants of removing OPN (No OPN), removing PRN (No PRN), and removing both (No OPN \& PRN).
\footnote{We ablate the OPN loss $\mathcal{L}^{\text{OPN}}_{\bm{\varphi}}$ or the PRN loss $\mathcal{L}^{\text{PRN}}_{\bm{\phi}}$ in Eq.~(\ref{loss}) when removing the OPN or the PRN module, respectively, and we ablate both losses when removing the two modules together.
During ablation, we keep the network architecture unchanged.
}
The results show that the performance of our method degenerates when removing either module of OPN and PRN, and degenerates more when removing both modules.
It successfully verifies that both the OPN and PRN modules boost the multi-agent coordination due to facilitating more accurate interactions between agents.

\subsection{StarCraft II Multi-Agent Challenge}

As benchmark testbeds, SMAC provides fully-cooperative battles for a group of agents learning to defeat the opponent team.
Typically, the game is framed as a competitive problem: an agent takes the role of a human player, making macromanagement decisions and performing micromanagement as a puppeteer that issues orders to individual units from a centralized controller.
In SMAC, each unit's actions are conditioned on local observations instead of the global game state.
Then, in several challenging combat scenarios, the group of these independent agents battles an opposing army under the centralized control of the build-in game AI. 
These challenges require agents to learn cooperative behaviours under partial observability, e.g., \emph{focus fire} and \emph{avoid overkill}. 
% In the carefully designed scenarios, there are two armies of units battling to defeat its enemy. The first army is controlled by the learned allied agents. The second army consists of enemy units controlled by the built-in game AI.
Each scenario consists of two armies of units battling to defeat each other, one controlled by the learner and the other controlled by the built-in game AI.
An episode ends when all units of either army have died or when a pre-defined step limit is reached. The goal is to maximize the win rate of the learned policies, i.e., the expected ratio of games won to games played.

Agents receive local observations drawn within their field of view, which is called \emph{sight range}. Agents can only observe other agents if they are both alive and located within the sight range. 
For agent $i$, the observation is formalized as
\begin{align}
	o^i_t =  \{ ( & \text{distance}_{i,j}, \Delta x_{i,j}, \Delta y_{i,j}, \nonumber\\
	& \text{health}_j, \text{sheild}_j, \text{unit type}_j ) \}_{\forall j\in \mathcal{U}},
\end{align}
where $j$ is a unit in the unit set $\mathcal{U}$ within the battle, $\text{distance}_{i,j}$ is the distance between units $i$ and $j$, and $\Delta x_{i,j}$ and $\Delta y_{i,j}$ are the offsets between $i$ and $j$ in axes $x$ and $y$, respectively. 
$\text{health}_j$, $\text{sheild}_j$, and $\text{unit type}_j$ denote the attributes of unit $j$. 
Specifically, if unit $j$ is not in the sight range of $i$, the above observation vectors related to unit $j$ are padded with $0$.

The available actions consist of move\text{[}direction\text{]} \text{(}four directions: north, south, east, and west\text{)}, attack\text{[}enemy\_id\text{]}, stop, and no\_op. Dead agents can only take no\_op action while alive agents cannot. Specifically, when unit $j$ is out of the \emph{shooting range} of unit $i$, the action of attacking $j$ is not available. We formalize the action space of agent $i$ as 
\begin{equation}
	a^i_t \!\in\!\left\lbrace  \text{no\_op}, \{\text{move[direction }d]\}, \{\text{attack unit }j\}_{j\in \mathcal{U} } \right\rbrace.
\end{equation}
% where $d$ is the direction in the set of $\{\text{north, south, east, west}\}$.

The reward contains three parts of the damage to enemies, killing enemies, and winning the battle as
\begin{align}
	r_t^i = & \sum\nolimits_{j\in{E}}  \left(\Delta \text{health}_j + \Delta \text{shield}_j\right)\nonumber \\
	& \quad + 10\cdot \Delta \text{\# dead enemies} + 200 \cdot \text{win flag},
\end{align}
where $\Delta \text{\# dead enemies}$ is the increased number of dead enemies, and $\text{win flag}$ is the flag of winning the combat.

\begin{figure*}[tb]\centering
	\subfigure[3m combat scenario.]{\label{3a}\includegraphics[width=0.48\columnwidth]{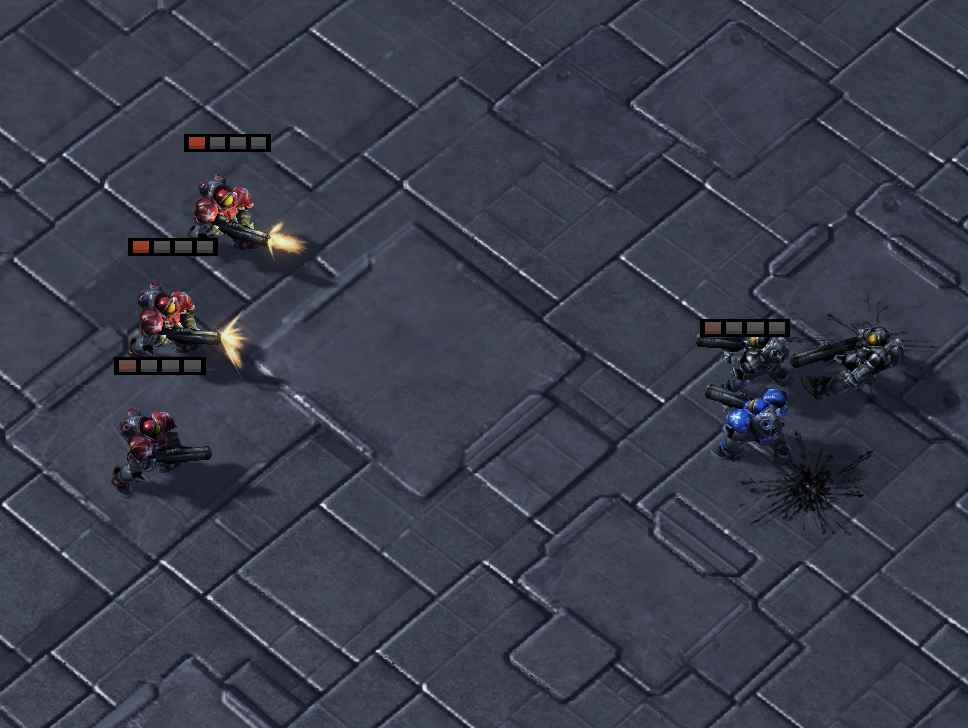}}
	\subfigure[8m combat scenario.]{\label{1b}\includegraphics[width=0.48\columnwidth]{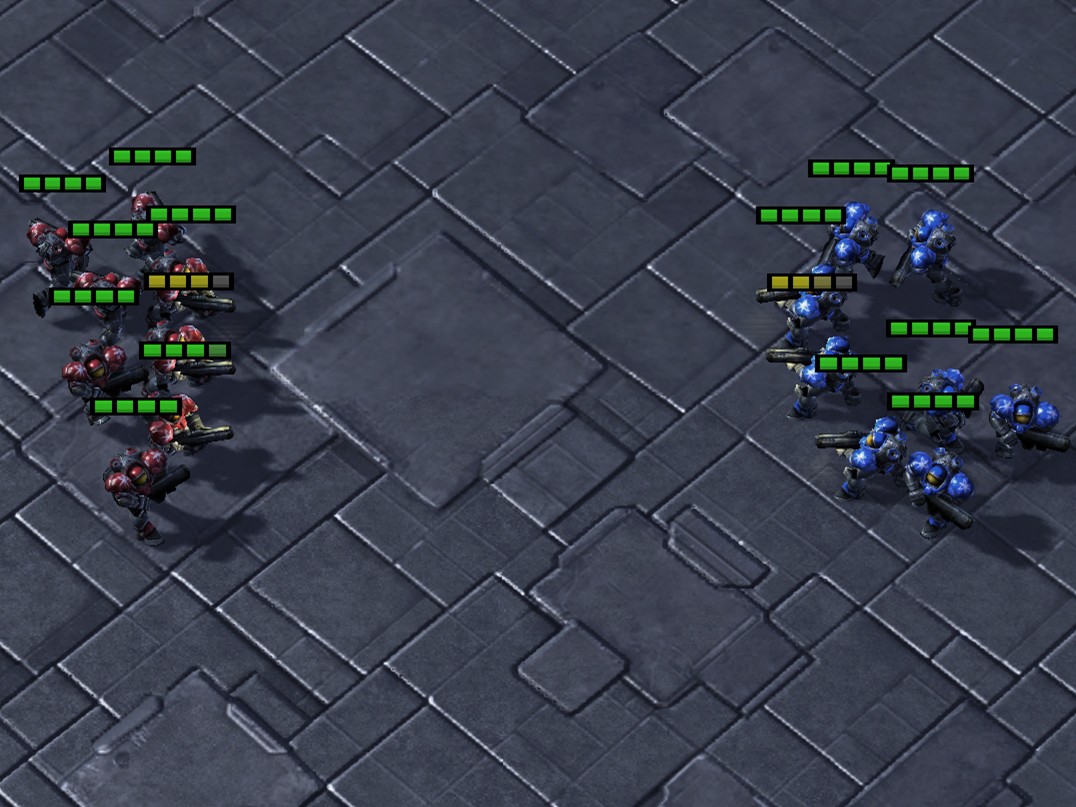}} 
	\subfigure[5m\_vs\_6m combat scenario.]{\label{3a}\includegraphics[width=0.48\columnwidth]{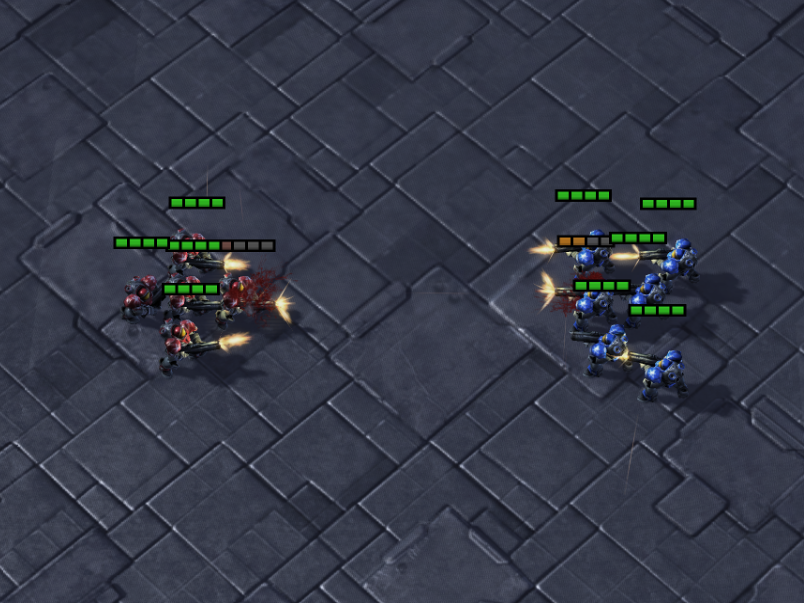}}
	\subfigure[2c\_vs\_64zg combat scenario.]{\label{3a}\includegraphics[width=0.48\columnwidth]{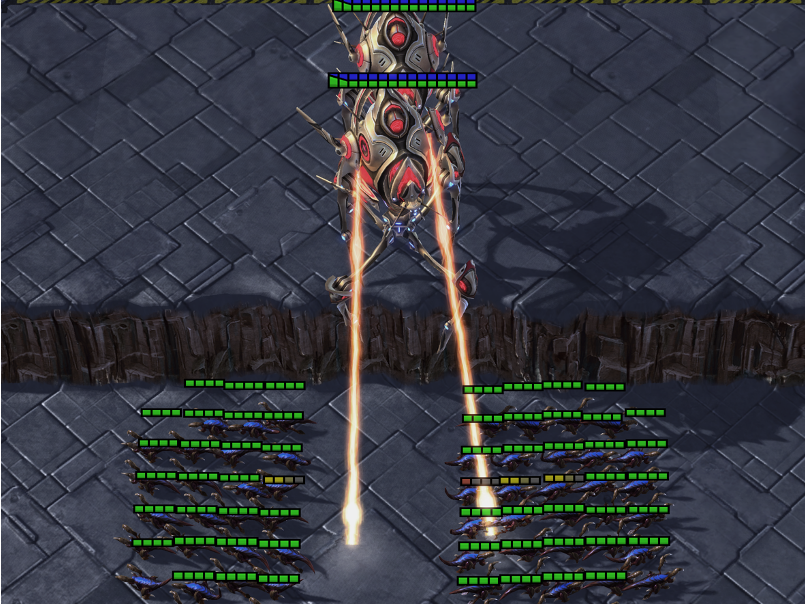}}
	\caption{The example screenshots of evaluation combat scenarios SMAC.}
	\label{smacmap}
\end{figure*}

\begin{table*}[tb]
	\centering
	\caption{The environment settings of the four evaluation combat scenarios in SMAC.}
	\begin{tabular}{c|cccc|cc}
		\toprule
		Scenarios		& Difficulty    &  Controlled agents    & Opponents   & Max steps & Observation size & Action size \\
		\midrule	
		3m		& very-hard		& 3 marines		& 3 marines		& 60 & 30 & 9  \\
		\midrule
		8m	    & very-hard       & 8 marines        & 8 marines       & 120  & 80 & 14  \\
		\midrule
		5m\_vs\_6m	    & very-hard       & 5 marines        & 6 marines       & 70  & 55 & 12  \\
		\midrule
		2c\_vs\_64zg	    & very-hard       & 2 colossi        & 64 zerglings       & 400  & 332 & 70  \\
		\bottomrule
	\end{tabular}
	\label{smacsettings}
\end{table*}

Since many QMIX-based approaches are verified to achieve good performance in fully-cooperative tasks, we take the following state-of-the-art MARL algorithms as baselines: QMIX, TarMAC+QMIX, and NDQ.
\footnote{NDQ itself is based on QMIX.}
We evaluate the methods on four combat maps: 3m, 8m, 5m\_vs\_6m, and 2c\_vs\_64zg.
In these maps, the neighboring relationship between agents is predefined according to their initial positions, and the convolution kernel of DCCP is with the same size as the kernel in ATSC. 
With this neighborhood formalization, we are capable of evaluating our method beyond the grid-world environments.
Fig.~\ref{smacmap} shows example screenshots of the four combat scenarios, and Table~\ref{smacsettings} presents the settings of the evaluation combat scenarios.
The difficulty of StarCraft II built-in AI is set to $\textit{very-hard}$ in all maps.
In each episode, positive rewards are given for the positive health point difference between the controlled agent team and the opponent, and otherwise, the reward is zero.
A large positive reward is given for winning the episode by eliminating the opponent, and otherwise, the reward is zero.
We evaluate the tested methods per $100$k steps during training, and in each evaluation, we run the game for $20$ episodes to calculate the win rate.
Table~\ref{smac_hyper} and Table~\ref{smac_net} present training hyperparameters and the network architecture, respectively.

\begin{table}[tb]
\setlength{\tabcolsep}{1.5mm}
\renewcommand\arraystretch{1.1}
	\centering
	\caption{Training hyperparameters of our method in SMAC.  $T$ is the number of maximum steps in each learning episode.}
	\begin{tabular}{c|cccccc}
		\toprule
		Task		& replay buffer size    &  batch size    & $\lambda_1$   & $\lambda_2$   & $\alpha$ & optimizer\\
		\midrule	
		SMAC	    & $32\times T$        & $32$        & $0.05$        & $0.01$ & $5$e-$4$  & Adam \\
		\bottomrule
	\end{tabular}
	\label{smac_hyper}
\end{table}

\begin{table*}[tb]
\renewcommand\arraystretch{1.1}
\setlength{\tabcolsep}{0.8mm}
	\centering
	\caption{The network architecture of our method in SMAC.}
	\begin{tabular}{c|c|ccccc|cc}
		\toprule
		\multirow{2}{*}{Scenarios} & 	\multirow{2}{*}{Modules}   & \multicolumn{5}{c|}{Encoder - layer size}   &\multicolumn{2}{c}{DCCP} \\
		& & input   &  hidden \#$1$ & hidden \#$2$ & LSTM size  & output  & \# kernels    & kernel size \\
		\midrule	
		\multirow{3}{*} {3m} 	& OPN   & $39$ ($30+9$) & \multirow{3}{*}{$256$}    &\multirow{3}{*}{$64$}	& $64$ 	& $30$  & \multirow{3}{*}{$10$} & \multirow{3}{*}{$3\times3$ }  \\
								& PRN	& $39$ ($30+9$)	&                           &                      	& $64$	& $9$   &                       &       \\
								& VFN	& $69$ ($30+30+9$) &                		&                   	& - 	& $9$	&                       &      \\
		\midrule			
		\multirow{3}{*} {8m}  	& OPN	& $94$ ($80+14$)    &  \multirow{3}{*}{$256$}	& \multirow{3}{*}{$64$}	&$64$	& $60$	&\multirow{3}{*}{$10$}  & \multirow{3}{*}{$5\times5$ }      \\
								& PRN	   		&      $94$ ($80+14$)  	& 		&	&$64$	& $14$			&           &       \\
								& VFN		    	&      $174$ ($80+80+14$) 	& 	&	&-	& $14$			&          &       \\
		\midrule			
		\multirow{3}{*} {5m\_vs\_6m}  	& OPN	    &      $67$ ($55+12$)  	&\multirow{3}{*}{$256$}		&\multirow{3}{*}{$64$}	&$64$	& $55$			&\multirow{3}{*}{$10$}           &  \multirow{3}{*}{$3\times3$ }       \\
								& PRN	   		&      $67$ ($55+12$)  	& 	& &$64$ 	& $12$			&          &        \\
								& VFN		    	&      $122$ ($55+55+12$) 	& 	&	&-	& $12$			&          &       \\
		\midrule			
		\multirow{3}{*} {2c\_vs\_64zg}  	& OPN	    &      $402$ ($332+70$)  	& \multirow{3}{*}{$256$}		&\multirow{3}{*}{$64$} &$64$		& $32$			&\multirow{3}{*}{$10$}           & \multirow{3}{*}{$3\times3$ }       \\
								& PRN	   		&       $402$ ($332+70$)   	& 	&	&$64$	& $70$			&        &     \\
								& VFN		    	&      $734$ ($332+332+70$) 	& 	&	&-	& $70$			&       &       \\
		
		\bottomrule
	\end{tabular}
	\label{smac_net}
\end{table*}

\begin{figure*}[tb]\centering
	\subfigure[3m, $\textit{very-hard}$]{\includegraphics[width=0.9\columnwidth]{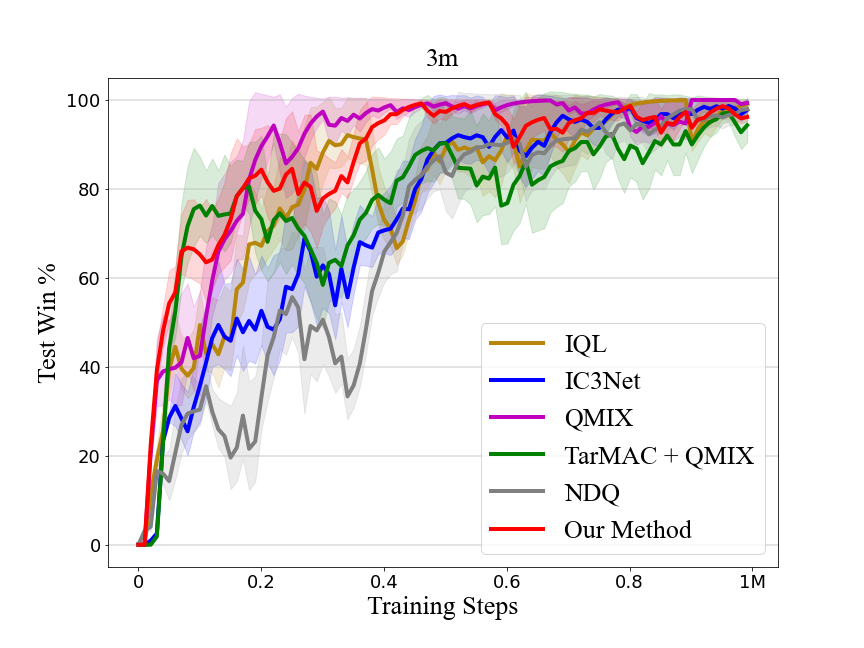}}
	\subfigure[8m, $\textit{very-hard}$]{\includegraphics[width=0.9\columnwidth]{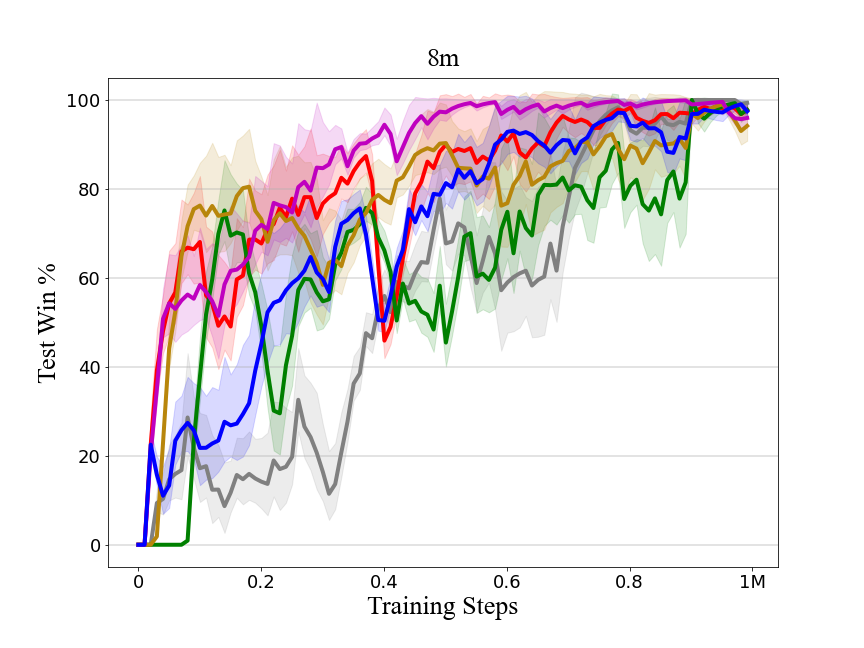}}
	\subfigure[5m\_vs\_6m, $\textit{very-hard}$]{\includegraphics[width=0.9\columnwidth]{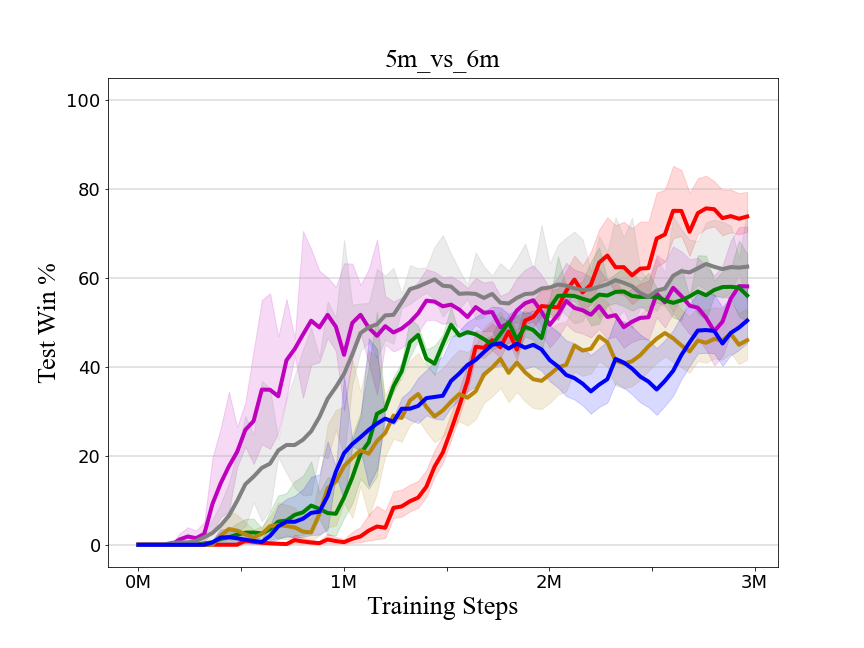}}
	\subfigure[2c\_vs\_64zg, $\textit{very-hard}$]{\includegraphics[width=0.9\columnwidth]{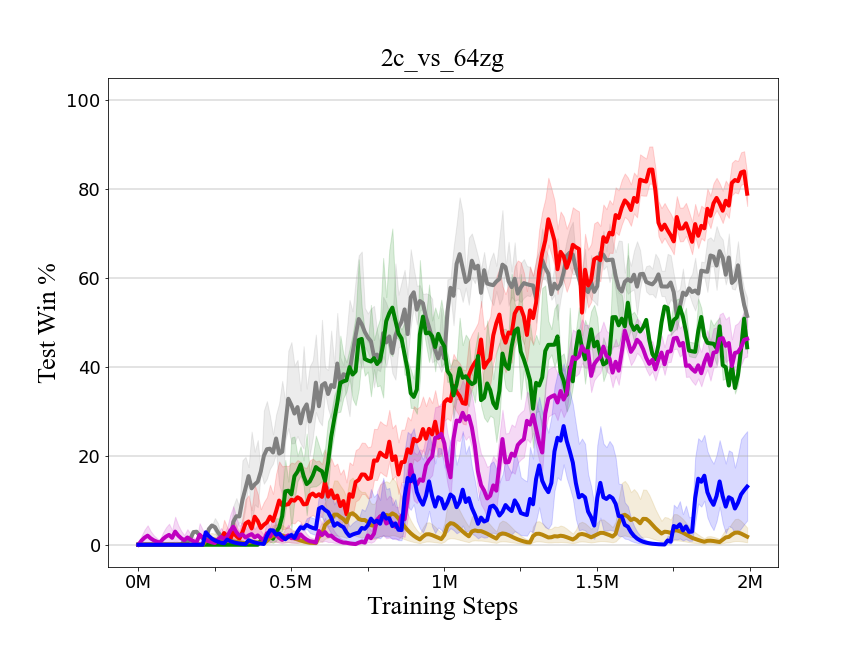}}
	\caption{Win rate during training in SMAC.}
	\label{fig5}
\end{figure*}

Fig.~\ref{fig5} presents evaluation curves of the win rate in the scenarios with $\textit{very-hard}$ built-in AIs, and Table~\ref{tab2} shows numerical results of evaluating the trained models. 
It is observed that our method generally achieves the highest and the most stable win rate in these combat maps.
The performance gap in terms of the final win rate is more pronounced in the complex combat maps of 5m\_vs\_6m and 2c\_vs\_64zg, which demonstrates the capability of our method for efficiently facilitating coordination across agents.
Moreover, our method learns the effect of joint actions without using the privileged state information from the environment, while SMAC can provide access to this information for methods like TarMAC. 
This makes our method applicable for a wider range of scenarios, as in many multi-agent tasks we do not have access to privileged full state information even during training.

\begin{table*}[tb]\renewcommand\arraystretch{1.1}
	\centering
	\caption{Evaluation win rate of the trained models in SMAC.}
	\begin{tabular}{l|rrrr}
		\toprule
		Methods			& 3m, hard			    & 8m, hard	            & 5m\_vs\_6m, hard        & 2c\_vs\_64zg, hard\\
		\midrule	
		IQL             & $99\pm 0.03 \%$       & $99\pm 0.69 \%$       & $48\pm 5.26 \%$       & $7\pm 4.91 \%$ \\		
		IC3Net         & $98\pm 0.93 \%$       & $99\pm 0.67 \%$       & $50\pm 5.01 \%$       & $27\pm 8.62 \%$ \\
		\midrule
		QMIX             & $\bm{100\pm 0.14} \%$       & $99\pm 0.06 \%$       & $58\pm 8.11 \%$       & $48\pm 3.11 \%$ \\
		TarMAC + QMIX          & $98\pm 1.99 \%$       & $ 100\pm 0.12 \%$       & $59\pm 2.50 \%$       & $54\pm 8.20 \%$ \\
		NDQ             & $98\pm 1.41 \%$       & $\bm{100\pm 0.08} \%$       & $63\pm 5.73 \%$       & $66\pm 2.17 \%$ \\
		\midrule
		Our Method     & $\bm{100\pm 0.38 \%}$  & $99\pm 0.93 \%$  & $\bm{76\pm 4.49} \%$       & $\bm{84\pm 3.18} \%$ \\
		\bottomrule
	\end{tabular}
	\label{tab2}
\end{table*}

\section{Conclusion}\label{conclusions}
In this paper, we propose a new MARL method based on local communication learning.
% We introduce two components to facilitate multi-agent coordination: one is exploiting the ability of depthwise convolution to efficiently learn local communication, and the other is enhancing the mean-field approximation by a supervised policy tracking module and a learnable compensation term to obtain a more accurate mean-field estimate.
We facilitate efficient coordination between neighboring agents by exploiting the ability of depthwise convolution to learn a local communication protocol, and by enhancing the mean-field approximation with a supervised policy rectification network (PRN) and a learnable compensation term.
Empirical results and an ablation study show that our method achieves efficient coordination and outperforms several baseline approaches on the ATSC and SMAC tasks.
Our future work will focus on learning more efficient communication protocols using graph structures and attention mechanisms.
Another insightful direction would be to develop efficient local communication protocols for more complex multi-agent systems where the neighborhood may vary with a dynamic communication topology.

% \clearpage
\footnotesize
\bibliographystyle{IEEEtranN}
\bibliography{reference}

\begin{IEEEbiography}[{\includegraphics[width=1.0in,height=1.25in,clip,keepaspectratio]{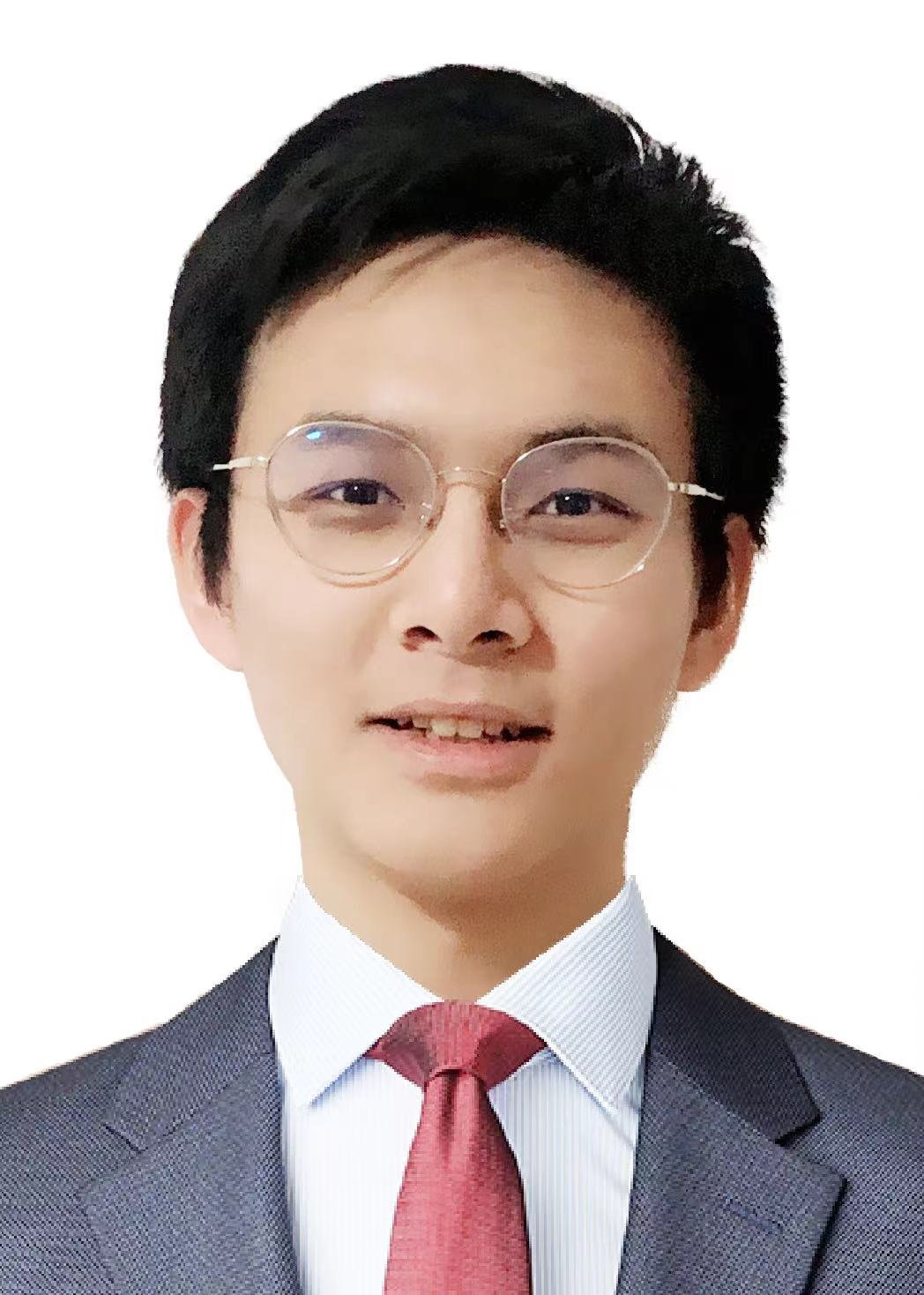}}]{Donghan Xie}
received the B.E. degree in mechanical engineering from the School of Mechanical Engineering, Shandong University, Jinan, China, in 2018, and the M.S. degree in the Department of Control Science and Intelligence Engineering, School of Management and Engineering, Nanjing University, Nanjing, China, in 2021.
His current research interests include multi-agent reinforcement learning and machine learning.
\end{IEEEbiography}

\begin{IEEEbiography}[{\includegraphics[width=1.0in,height=1.25in,clip,keepaspectratio]{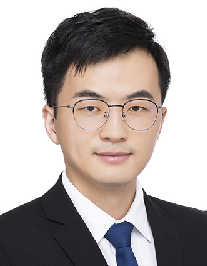}}]{Zhi Wang} (S'19-M'20) received the Ph.D. degree in machine learning from the Department of Systems Engineering and Engineering Management, City University of Hong Kong, Hong Kong, China, in 2019, and the B.E. degree in automation from Nanjing University, Nanjing, China, in 2015. He is currently an Associate Research Fellow with the Department of Control Science and Intelligence Engineering, School of Management and Engineering, Nanjing University, Nanjing, China. He holds visiting positions at the University of New South Wales, Australia and the State Key Laboratory of Management and Control for Complex Systems, Institute of Automation, Chinese Academy of Sciences, China.
	
His current research interests include reinforcement learning, machine learning, and robotics.
He served as the Associate Editor for IEEE International Conference on Systems, Man, and Cybernetics 2021 and 2022, and IEEE International Conference on Networking, Sensing, and Control 2020.
\end{IEEEbiography}

\begin{IEEEbiography}[{\includegraphics[width=1.0in,height=1.25in,clip,keepaspectratio]{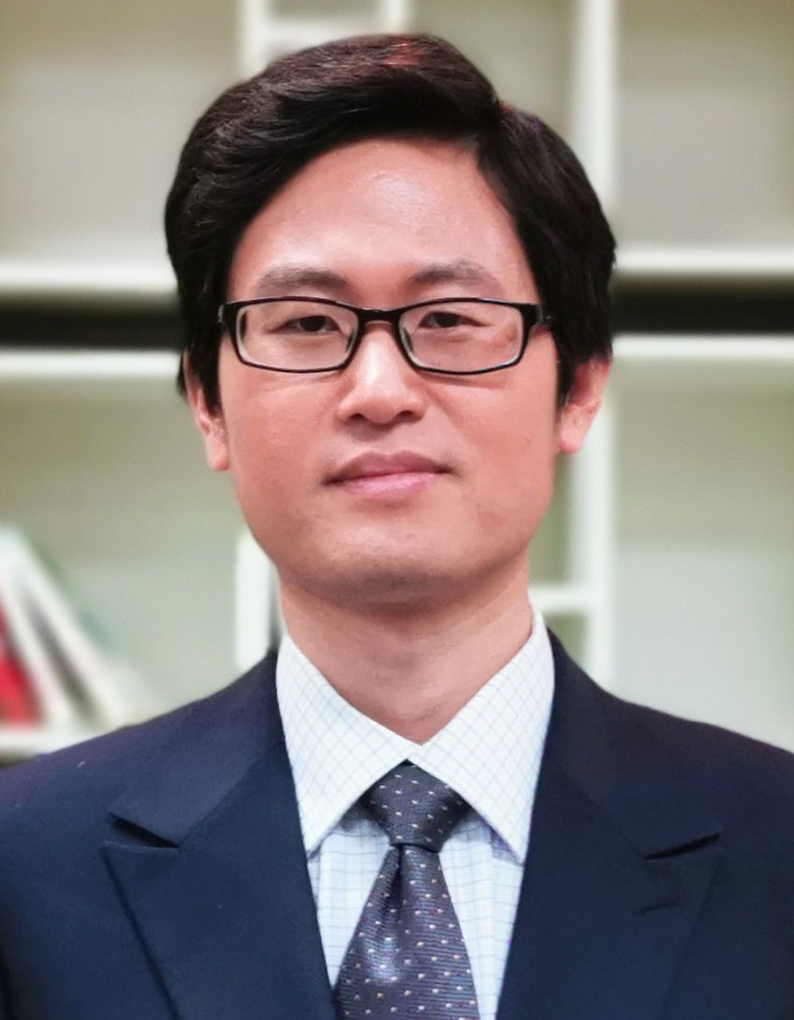}}]{Chunlin Chen}
(S'05-M'06-SM'21) received the B.E. degree in automatic control and Ph.D. degree in control science and engineering from the University of Science and Technology of China, Hefei, China, in 2001 and 2006, respectively. 
He is currently a full professor and the vice dean of School of Management and Engineering, Nanjing University, Nanjing, China. 
He was a visiting scholar at Princeton University, Princeton, USA, from 2012 to 2013. He had visiting positions at the University of New South Wales, Canberra, Australia, and the City University of Hong Kong, Hong Kong, China.

His recent research interests include reinforcement learning, mobile robotics, and quantum control. 
He is the Chair of Technical Committee on Quantum Cybernetics, IEEE Systems, Man and Cybernetics Society.
\end{IEEEbiography}

\begin{IEEEbiography}[{\includegraphics[width=1.0in,height=1.25in,clip,keepaspectratio]{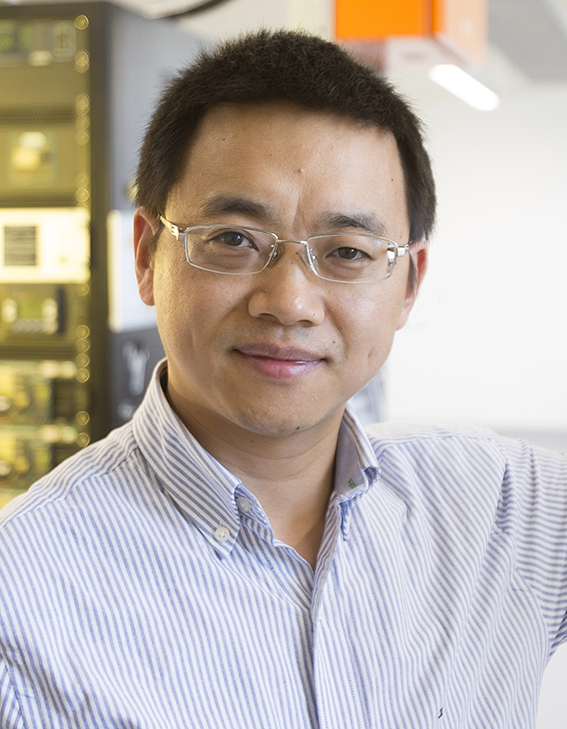}}]{Daoyi Dong}
received the B.E. degree in automatic control and the Ph.D. degree in engineering from the University of Science and Technology of China, Hefei, China, in 2001 and 2006, respectively.

He was an Alexander von Humboldt Fellow at AKS, University of Duisburg-Essen, Duisburg, Germany.
He was with the Institute of Systems Science, Chinese Academy of Sciences, Beijing, China, and with Zhejiang University, Hangzhou, China. He had visiting positions at Princeton University, NJ, USA; RIKEN, Wako-Shi, Japan; and The University of Hong Kong, Hong Kong. 
He is currently a Scientia Associate Professor at the University of New South Wales, Canberra, ACT, Australia. His research interests include quantum control and machine learning.

Dr. Dong was awarded the ACA Temasek Young Educator Award by the Asian Control Association and was a recipient of Future Fellowship, the International Collaboration Award and the Australian Post-Doctoral Fellowship from the Australian Research Council, and a Humboldt Research Fellowship from the Alexander von Humboldt Foundation of Germany. 
He is a Member-at-Large, Board of Governors, and was the Associate Vice President for Conferences and Meetings, IEEE Systems, Man and Cybernetics Society. He served as an Associate Editor for the IEEE TRANSACTIONSON NEURAL NETWORKS AND LEARNING SYSTEMS from 2015 to 2021. He is currently an Associate Editor of the IEEE TRANSACTIONS ON CYBERNETICS and a Technical Editor of the IEEE/ASME TRANSACTIONS ON MECHATRONICS. 
He is a Fellow of the IEEE.
\end{IEEEbiography}

\end{document}